\documentclass[12pt,a4paper]{article}
\renewcommand{\baselinestretch}{1.2} 
\usepackage[pdftex]{graphicx}
\usepackage[cmex10]{amsmath}
\usepackage{amsfonts}
\usepackage{amssymb}
\usepackage{amsthm}
\newtheorem{mydef}{Definition}

\usepackage{algorithm}
\usepackage{algpseudocode}
\usepackage{cite}
\usepackage{color}
\usepackage{authblk}
\usepackage{enumerate}
\usepackage{subfigure}
\usepackage[font=footnotesize ]{caption}
\usepackage{anysize}
\marginsize{2cm}{2cm}{2cm}{2cm}

\title{Metaheuristic Design of Feedforward Neural Networks: A Review of Two Decades of Research}
\author[1]{Varun~Kumar~Ojha\thanks{Corresponding Author~\\Engineering Applications of Artificial Intelligence 60 (2017) 97--116}}
\author[2]{Ajith~Abraham}
\author[1]{V\'{a}clav~Sn\'{a}\v{s}el}
\affil[1]{Dept. of Computer Science, V{\v{S}}B-Technical University of Ostrava, Ostrava, Czech Republic}
\affil[2]{Machine Intelligence Research Labs (MIR Labs), Auburn, WA, USA}

\date{}
\begin{document}
\maketitle
\begin{abstract}
Over the past two decades, the feedforward neural network  (FNN) optimization has been a key interest among the researchers and practitioners of multiple disciplines. The FNN optimization is often viewed from the various perspectives: the optimization of weights, network architecture, activation nodes, learning parameters, learning environment, etc. Researchers adopted such different viewpoints mainly to improve the FNN's generalization ability. The gradient-descent algorithm such as backpropagation has been widely applied to optimize the FNNs. Its success is evident from the FNN's application to numerous real-world problems. However, due to the limitations of the gradient-based optimization methods, the metaheuristic algorithms including the evolutionary algorithms, swarm intelligence, etc., are still being widely explored by the researchers aiming to obtain generalized FNN for a given problem. This article attempts to summarize a broad spectrum of FNN optimization methodologies including conventional and metaheuristic approaches. This article also tries to connect various research directions emerged out of the FNN optimization practices, such as evolving neural network (NN), cooperative coevolution NN, complex-valued NN, deep learning, extreme learning machine, quantum NN, etc. Additionally, it provides interesting research challenges for future research to cope-up with the present information processing era.~
\end{abstract}

\clearpage

\section{Introduction}
\label{sec_introSurvey}
Back in 1943 McCulloch and Pitts~\cite{annMcCollucPitts1943} proposed a computational model inspired by the human brain, which initiated the research on artificial neural network (ANN). ANNs are capable of learning and recognizing and can solve a broad range of complex problems. Feedforward neural networks (FNNs) are the special type of ANN models. The structural representation of an FNN makes it appealing because it allows perceiving a computational model (a function) in a structural/network form. Moreover, it is the structure of an FNN that makes it a universal function approximator, which has the capabilities of approximating any continuous function~\cite{annHornik1991approximation}. Therefore, a wide range of problems is solved by the FNNs, such as pattern recognition~\cite{annAppPR2}, clustering and classification~\cite{annAppClass}, function approximation~\cite{annAppFunAprox}, control~\cite{annAppControl}, bioinformatics~\cite{annAppBio}, signal processing~\cite{annAppSignal}, speech processing~\cite{annAppSpeech}, etc.

The structure of an FNN consists of several neurons (processing units) arranged in layer-by-layer basis and the neurons in a layer have connections (weights) from the neurons at its previous layer. Fundamentally, an FNN optimization/learning/training is met by searching an appropriate network structure (a function) and the weights (the parameters of the function)~\cite{haykin2009neural}. Finding a suitable network structure includes the determination of the appropriate neurons (i.e., activation functions), the number of neurons, and the arrangements of neurons, etc. Similarly, finding the weights indicates the optimization of a vector representing the weights of an FNN. Therefore, learning is an essential and distinguished aspect of the FNNs. 

Numerous algorithms, techniques, and procedures were proposed in the past for the FNNs optimization. Earlier, in FNN research, only the gradient-based optimization techniques were the popular choices. However, gradually because of the limitations of gradient-based algorithms, the necessity of metaheuristic-based optimization methods were recognized. 

Metaheuristics formulate the FNN components, such as weights, structure, nodes, etc., into an optimization problem. Metaheuristics implement various heuristics for finding a near-optimum solution. Additionally, a multiobjective metaheuristic deals with the multiple objectives simultaneously. The existence of multiple objectives in the FNNs optimization is evident since the minimization of FNN's approximation error is desirable at one hand, and the generalization and model's simplification is at the other. 

In a metaheuristic or multiobjective metaheuristic treatment to an FNN, an initial population of FNNs is guided towards a final population, where usually the best FNN is selected. However, selecting only the best FNN from a population may not always produce a general solution. Therefore, to achieve a general solution without any significant additional cost, an ensemble of many candidates chosen from a metaheuristic final population is recommended. 

This article provides a comprehensive literature review to address the various aspects of the FNN optimization, such as:
\begin{enumerate}
\item The importance of an FNN as a function approximator and its preliminary concepts (Section~\ref{sec_annAnn}), including the introduction to the factors influencing FNN optimization (Section~\ref{sub_FNNoptIssues}) and introduction to the conventional optimization algorithms (Section~\ref{sub_annGradOpt}).
\item The role of metaheuristics and hybrid metaheuristics in FNNs optimization (Section~\ref{sec_fnnMetaOpt}).
\item The role of multiobjective metaheuristics (Section~\ref{sec_fnnEMO}) and the ensemble methods (Section~\ref{sec_ensemble}). 
\item The current challenges and future research directions (Section~\ref{sec_challanges}).   	
\end{enumerate}

\section{Feedforward neural networks}  
\label{sec_annAnn}
The intelligence of human brain is due to its massively parallel neurons network system. In other words, the architecture of the brain. Similarly, a proper design of an ANN offers a significant improvement to a learning system. The components, such as nodes, weights, and layers are responsible for the developments of various ANN models. 

A single layer perceptron (SLP) consists of an input and an output layer, and it is the simplest form of ANN model~\cite{annRosenblatt1958,annTutorialJain1996}. However, SLPs are incapable of solving nonlinearly separable patterns~\cite{annMinsky1969}. Hence, a multilayer perceptron (MLP) was proposed, which addressed the limitations of SLPs by including one or more hidden layers in between an input and an output layer~\cite{annBpWerbos1975}. Initially, the backpropagation (BP) algorithm was used for the MLP training~\cite{annBpRumelhart1986}. A trained MLP is then found capable of solving nonlinearly separable patterns~\cite{annBpRumelhart1986}. In fact, MLPs (in general FNNs) are capable of addressing a large class of problem pertaining to pattern recognition and prediction. Moreover, an FNN is considered as a {\textbf{universal approximator}}~\cite{annHornik1991approximation}. Cybenko~\cite{annCybenko1989} referring to Kolmogorov's theorem\footnote{Kolmogorov's theorem: ``All continuous functions of $n$ variables have an exact representation in terms of finite superpositions and compositions of a small number of functions of one variable~\cite{annKolmogorov1957}."} showed that an FNN with only a single internal hidden layer---containing a finite number of neurons with any continuous sigmoidal nonlinear activation function---can approximate any continuous function. Also, the FNN structure (architecture) is itself capable enough to be a universal approximator~\cite{annHornik1989multilayer,annHornik1991approximation}. Hence, several researchers praised FNN for its universal approximation ability~\cite{annKuurkova1992kolmogorov,annLeshno1993multilayer,annHuang1998upper,annHuang2006universal}. 

Many other ANN models, like radial basis function~\cite{annRbfLowe1988Mvar} and support vector machine~\cite{annSvmCorinna1995} are a special class of three-layer FNNs. They are capable of solving regression and classification problems using supervised learning methods. In contrast, adaptive resonance theory~\cite{annArtGrossberg1987competitive}, Kohenen's self-organizing map~\cite{annSomKohonen1982}, and learning-vector-quantization~\cite{annSomKohonen1982} are two-layer FNNs that are capable of solving pattern recognition and data compression problems using unsupervised learning methods. 

Additionally, the ANN architecture with feedback connections, in other words, a network where connections between the nodes may form cycles is known as a \textbf{recurrent neural network} (RNN) or feedback network model. The RNNs are good at performing sequence recognition/reproduction or temporal association/prediction tasks. RNNs such as Hopfield network~\cite{annHopfield1982neural} and Boltzmann machine~\cite{annAckley1985} are good at the application for memory storage and remembering input--output relations. Moreover, Hopfield network was designed for solving nonlinear dynamic systems, where the stability of a dynamic system is studied under the neurodynamic paradigm~\cite{annHopfield1982neural}. 

A collection of RNN models, such as temporal RNN~\cite{annDominey95}, echo state RNN~\cite{annJaeger2001echo}, liquid state machine~\cite{annNatschlager2002} and backpropagation de-correlation~\cite{annSteil04} forms a paradigm called reservoir computing, which addresses several engineering applications including nonlinear signal processing and control. Although some other ANN models that are capable of doing a similar task that of the FNNs were pointed out in this Section, the discussion in this article is; however, limited to only FNNs.
\subsection{Components of FNNs}
\label{sec_fnn_components}
\begin{figure}
	\centering
	\subfigure[Three-layer feedforward neural network]{
	\includegraphics[width=0.4\textwidth]{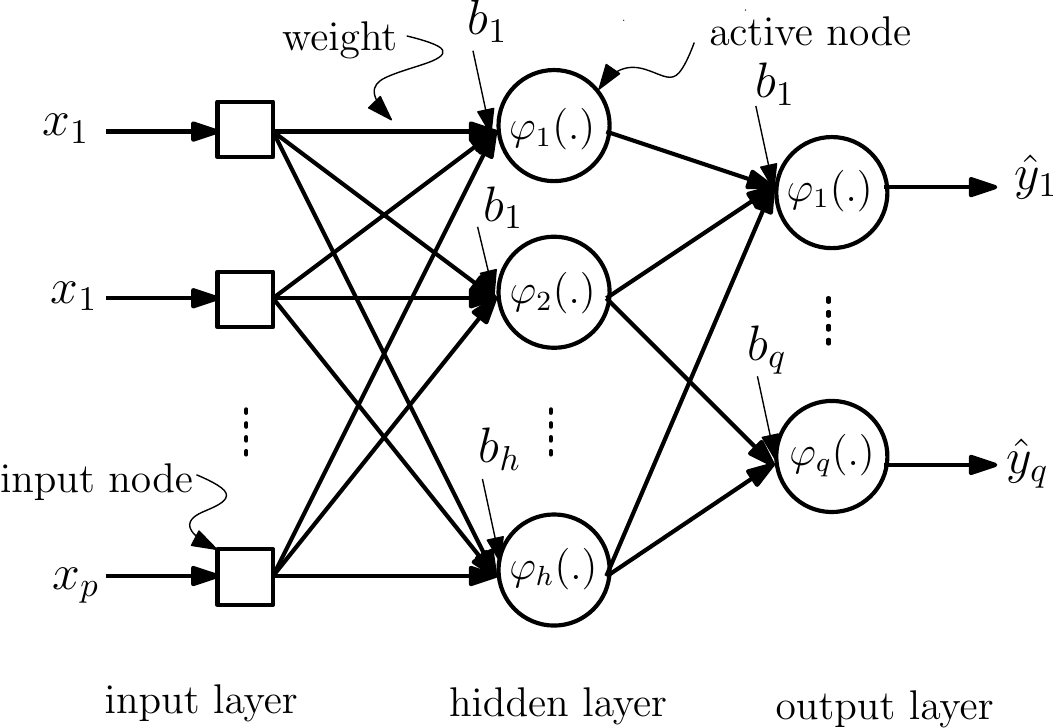}
	\label{fig_genFNN}
     }
     \subfigure[Node of the network]{
     \includegraphics[width=0.4\textwidth]{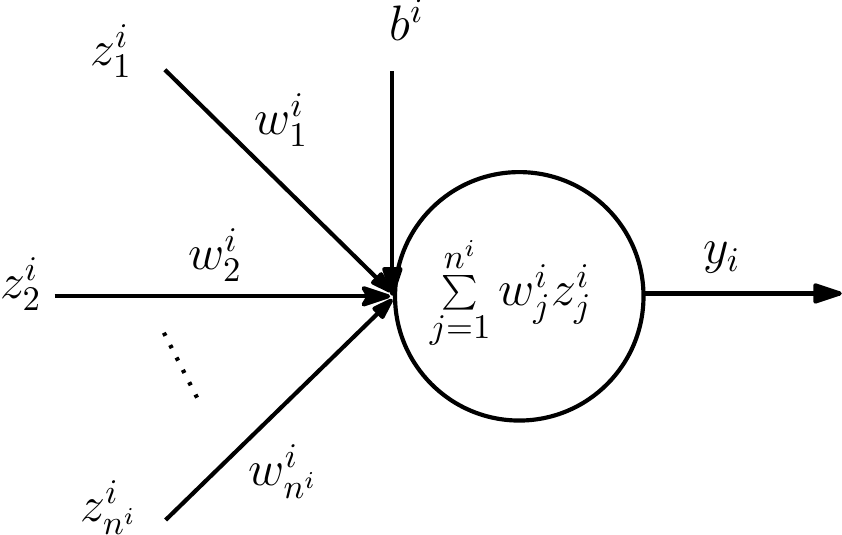}
     \label{fig_genFNN_node}
     }
	\caption{Three-layer feedforward neural network (a), where input layer has $ p $ input nodes, hidden layer has $ h $ activation functions, and output layer has $ q $ nodes.}
	\label{fig_genFNN_fig}
\end{figure}

FNNs are the computational models that consist of many neurons (\textit{node}), which are connected using synaptic links (\textit{weights}) and are arranged in layer-by-layer basis. Thus, the FNNs have a specific structural configuration (\textit{architecture}) in which the nodes at a layer have forward connections from the nodes at its previous layer (Fig.~\ref{fig_genFNN}). 
A node of an FNN is capable of processing information coming through the connection weights (Fig.~\ref{fig_genFNN_node}). Mathematically, the output $ y_i $ (excitation) of a node (node indicated as $ i $) is computed as: 
\begin{equation}
	\label{eq_yAnn}
	y_i = \varphi_i \left( \sum\limits_{j = 1}^{n^i} w^i_{j}z^i_j + b^i \right), 
\end{equation}
where $ n^i $ is the total incoming connections, $z^i$ is the input, $w^i$ is the weight,  $b^i$ is the bias, and $\varphi_i(\cdot)$ is the \textit{activation function} at the $ i $-th node to limits the amplitude of the output the node into a certain range.

Fig.~\ref{fig_genFNN} is a structural representation of an FNN, i.e., a phenotype of a function $ f(\text{\bf{x}},\text{\bf{w}}) $, which is parameterized by a $ p $-dimensional input vector $ \text{\bf{x}} = \langle x_1, x_2,\ldots, x_p \rangle$ and an $ n $-dimensional real-valued weight vector $\text{\bf{w}} = \langle w_1,w_2,\ldots,w_n \rangle$. The function $ f(\text{\bf{x}},\text{\bf{w}}) $ is a solution of a problem. Therefore, two tasks involved in solving a problem using an FNN are: to discover an appropriate function $ f(\text{\bf{x}},\text{\bf{w}}) $ (i.e., the architecture optimization) and to discover an appropriate weight vector $ \text{\bf{w}} $ (i.e., the weights optimization) using some \textit{learning algorithm}. 

The architecture optimization indicates the search for the appropriate activation functions at the nodes, the number of nodes, number of layers, the arrangements of the nodes, etc. Therefore, several components of an FNN optimization are: 
	 the connection \textbf{weights};
	 the \textbf{architecture} (number of layers in a network, 
	 the number of nodes at the hidden layers,
	 the arrangement of the connections between nodes);
	 the \textbf{nodes} (activation functions at the nodes); 
	 the \textbf{learning algorithms} (algorithms training parameters); and
	 the \textbf{learning environment}.
However, traditionally, the only component that was optimized was the weights of the connections by keeping other components fixed to the initial choice.

\subsection{Influencing factors in FNN optimization}
\label{sub_FNNoptIssues}

\subsubsection{Learning environments}
\label{sub_learningEnv}
An FNN is trained by supplying the training data ($ X, Y $) of $ N $ input--output pairs, i.e.,  $ X = (\text{\textbf{x}}_1, \text{\textbf{x}}_2, \ldots,\text{\textbf{x}}_N) $ and $ Y = (\text{\textbf{y}}_1, \text{\textbf{y}}_2, \ldots,\text{\textbf{y}}_N) $. Each input~$\text{\bf{x}}_i = \langle x_{i1},x_{i2},\ldots,x_{ip} \rangle$ is a $ p $-dimensional vector, and it has a corresponding $ q $-dimensional desired output vector $\text{\bf{y}}_i = \langle y_{i1}, y_{i2},\ldots, y_{iq} \rangle$. For the training data ($ X, Y $), an FNN produces an output $ \hat{Y} = (\hat{\text{\bf{y}}}_1, \hat{\text{\bf{y}}}_2, \ldots, \hat{\text{\bf{y}}}_N) $, where a vector $ \hat{\text{\bf{y}}}_i = \langle \hat{y}_{i1}, \hat{y}_{i2},\ldots, \hat{y}_{iq} \rangle$  is a $ q $-dimensional FNNs output, which is then compared with the desired output $ \text{\bf{y}}_i $, for all $ i = 1$ to $ N $ by using some error/distance/cost function. The minimization/reduction of the error/distance function, in an iterative manner, is referred as a \textbf{supervised learning}. One very commonly known supervised learning algorithm is Delta rule or Widrow-Hoff rule~\cite{annDeltaWidrow1960,annDeltaWidrow1959} in which the $ n $-dimensional weight vector $ \text{\textbf{w}} $ of an FNN is optimized as:
\begin{equation}
\label{eq_update}
\text{\bf{w}}^{t+1} = \text{\bf{w}}^t + \Delta \text{\bf{w}}^t,
\end{equation}
where $\Delta \mathrm{w}^t$ is weight change (an additive term) at $t$-th iteration. The weight change $\Delta \mathrm{w}^t$ is computed as:
\begin{equation}
\label{eq_changeInWeight}
\Delta \text{\bf{w}}^t_i = \eta^t e^t_i \text{\textbf{x}}^t_i,
\end{equation} 
where $\eta^t$ is a learning rate, which controls the magnitude of weight change at $ t $-th iteration and $e^t_i$ is the error at $ t $-th learning iteration corresponding to $ i $-th training input $ \text{\textbf{x}}^t_i $ presented to an FNN. The error $e^t_i$ at the $ t $-th iteration may be computed as: $ e^t_i = \sum_{j=1}^{q} (y^t_{ij} - \hat{y}^t_{ij})^2 $, where $ y^t_{ij} $ and $ \hat{y}^t_{ij} $ are the desired output and FNN's output at $ t $-th iteration respectively. 

Contrary to the supervised learning paradigm, there are two other learning forms for the spacial cases of FNNs: 1) the \textit{unsupervised learning}---for the unlabeled training data~\cite{annCompetitiveLearRumelhart1985}, and 2)  the \textit{reinforcement learning}---for the training data with insufficient input--output relations~\cite{annKaelbling96reinforcementlearning}. The focus of this article is, however, on supervised learning paradigms only.

\subsubsection{Error functions}
A supervised learning, essentially, is the minimization of the difference/distance between the desired output $ \text{\bf{y}}_i $ and the model's output $ \hat{\text{\bf{y}}}_i = f(\text{\bf{x}},\text{\bf{w}}) $ by comparing the difference/distance using a cost function~$c_f: Y \times \hat{Y} \longrightarrow \mathbb{R}_{\ge 0} $. For this propose, several cost function can be designed. For instance, in regression problems, \textit{mean squared error} is one of the commonly used cost function, which is written as:  
\begin{equation}
\label{eq_mse}
c_f(\text{\bf{y}}_i,\hat{\text{\bf{y}}}_i)  = \frac{1}{N}\sum\limits_{i = 1}^{N} \sum\limits_{j = 1}^{q} \left(y_{ij} - \hat{y}_{ij} \right)^2, 
\end{equation} 
where $ y_{ij} $ are the desired response and $ \hat{y}_{ij} $ are the FNN's responses, and their differences were summed over $N$ data pairs. Some other functions like sum of squared error, root of mean square error, mean absolute error, correlation coefficient, etc., can be used for evaluating the FNN's predictability~\cite{pearce2000evaluating}. 

The cost function~\eqref{eq_mse} or any similar squared-error-based cost function is inconsistent for solving classification problems~\cite{annPerfTwomey1995}. Instead, the \textit{percentage of good classification}, which has consistent behavior, can be used~\cite{annPerfTwomey1995}. However, the percentage of good classification is satisfactory until no preference was given to a particular class. Therefore, \textit{accuracy} and \textit{miss-classification rate} are used as the cost functions. A detailed list of the cost function for evaluating the classification problems is available in~\cite{pencina2008evaluating,annPerfSokolova2009systematic,classFernandez2010}. 

In this article, cost function mentioned for FNN optimization is discussed in a general sense, which can be thought as the equivalent to any other user-defined cost function. Another factor related to fitness of an FNN is to compare the cost functions of two or more FNN models~\cite{baranyi1999validating,van1994comparing}. Some researchers also argue to statistically compare the predicted outputs of two or more FNN models to establish the differences in their performances~\cite{diebold1995comparing}.

\subsubsection{Local minima problem}
\label{subsub_localMinima}
Let $c_f: S \longrightarrow \mathbb{R}_{\ge 0} $, where $S \subset \mathbb{R}^n$ is nonempty and compact (for detailed information about topological compactness, see~\cite{annSimovici2008mathematical}). Therefore, the following may be defined:
\begin{mydef}
	\label{def_globalOpt}
	A point $\text{\bf{w}}^* \in S$ is called \textbf{global minima} if $	c_f(\text{\bf{w}}^*) \leq c_f(\text{\bf{w}}) \text{ for any } \text{\bf{w}} \in S$ holds. 
\end{mydef} 
\begin{mydef}
	\label{def_localOpt}
	A point $\text{\bf{w}}^* \in S$ is called \textbf{local minima} if there exists $\epsilon > 0$, and an $\epsilon$-neighborhood $B_{\epsilon}(\text{\bf{w}}^*,\epsilon)$ around $\text{\bf{w}}^*$ such that $c_f(\text{\bf{w}}^*) \leq c_f(\text{\bf{w}}) \text{ for any } \text{\bf{w}} \in S \cap B_{\epsilon}(\text{\bf{w}}^*,\epsilon)$ holds.
\end{mydef}
Learning algorithms when to using the cost function~\eqref{eq_mse} or any similar function for FNN optimization has the tendency to fall in local minima~\cite{ensmblHansen1990}. Moreover, the geometrical structure (parameter space) of a three-layer perceptron may fall to local minima and plateaus during its optimization. It indicates that the critical point corresponding to global minima of a smaller FNN model (model with $h-1$ hidden units) can be a local or saddle point of a larger FNN model (model with $h$ hidden units)~\cite{annMinimaFukumizu1999local}. However, there are some ways to avoid or eliminate local minima in FNN optimization~\cite{annMininmaWessels1992avoiding,annMinimaToh2003deterministic}: 
\begin{enumerate}[1)]
	\item If the weights and training patterns are assigned randomly to a three-layer FNN that contains $ h $ neurons at the hidden layer, then a gradient-descent algorithm can avoid trapping into local minima~\cite{annMinimaPoston1991}.
	\item If linearly-separable training data and pyramidal network structure are taken, then the error surface will be local minima free~\cite{annMinimaGori1992problem}.
	\item If there are $ N $ many noncoincident input patterns to learn and three-layered FNN with $ N - 1 $ sigmoid hidden neurons and one dummy hidden neuron is used, then the corresponding error surface will be local minima free. 
	\item If the training algorithms can be improved as similar to as the global descent learning algorithm proposed by Cetin et al.~\cite{annMinimaCetin1993global} to replace gradient-descent algorithms, then it can avoid local minima.
\end{enumerate}
These four methods depend on the number of hidden neurons, the number of training samples, the number of output neurons, and a condition that says the number of hidden neurons should not be less than the number of training samples. Moreover, it does not necessarily guarantee to converge to global minima and to set preconditions for the number of hidden neurons and linearly separable training patterns are unlikely conditions for the real-world problems~\cite{annMinimaHuang1998local}.  

\subsubsection{Generalization} 
\label{sec_FNN_gen}
The generalization is a crucial aspect of an FNN optimization, where it is an ability to offer the general solutions rather than performing best for the particular cases. To achieve generalization, FNNs need to avoid both \textit{underfitting} and \textit{overfitting} during training, which is associated with high statistical \textit{bias} and high statistical \textit{variance}~\cite{annGenGeman1992neural}. Therefore, one has to address trade-offs between bias and variance. Also, for a good generalization, the number of training pattern should be sufficiently larger than the total number of connections in FNN~\cite{widrow199030}. 

The standard methods to achieving generalization are determining an optimum number of free parameters (i.e., equivalent to find an optimum network architecture), \textit{early stopping} of training algorithms, \textit{adding regularization term} with the cost function~\cite{eannGirosi1995regularization,bishop1995training}, and \textit{adding noise} to the training data.  

In \textit{early stopping}, a dataset is divided into three sets: a training set, cross-validation set, and test set. The early stopping scheme suggests stopping of training at the point (epoch) from which onward the cost function value computed on cross-validation set starts to rise~\cite{ensmblHansen1990,annGenAmari1997,prechelt1998automatic,annGenYao2007early}. Similarly, adding noise (jitters) into the training pattern improves FNN's \textit{generalization} ability and removing insignificant weights from a trained FNN improves its \textit{fault tolerance} ability~\cite{murray1994enhanced}. Moreover, generalization is related to sparsity and stability of a learning algorithm~\cite{bousquet2002stability}.

Now, if the approximation error of two FNN models trained on the same training data is close/similar, then the model with simple network structure (lower number of free parameters) should be selected as the best model. It is because the model with lower network complexity possesses higher generalization ability than the models with higher network complexity~\cite{reed1995similarities}. Moreover, the network with lower weight magnitude possesses better generalization ability~\cite{reed1995similarities}.

\subsection{Conventional optimization approaches}
\label{sub_annGradOpt}
Finding a suitable algorithm for the FNNs optimization has always been a difficult task. The FNN optimization using conventional gradient based algorithms is viewed as an unconstrained optimization problem~\cite{haykin2009neural,annLippamann1987}. The cost function $c_f$ has to be optimized to satisfy Definition~\ref{def_globalOpt}.
Therefore, the gradient of error $g^t$ at $ t $-th iteration is computed as:
\begin{equation}
\label{eq_gradient}
g^t = \frac{\partial c_f}{\partial \text{\bf{w}}^t} ,
\end{equation}
where $ g^t$ is a \textit{first-order partial derivative} of the cost function $c_f$ with respect to weight vector $\text{\bf{w}}$. Hence, a gradient-descent approach starts with an initial guess $\text{\bf{w}}_0$ and generates a sequence of weight vector $\text{\bf{w}}_1,\text{\bf{w}}_2, \ldots$ such that $c_f$ reduces in each iteration. The connection weights at iteration $ t $ are updated as: 
\begin{equation}
\label{eq_stepestDescent}
\text{\bf{w}}^{t+1} = \text{\bf{w}}^t + \Delta \text{\bf{w}}^t,
\end{equation}
where the weight change $\Delta \text{\bf{w}}^t \text{ is equal to } -\eta^t g^t$, and $ \eta^t $ is the learning rate. The weights updated using~\eqref{eq_gradient} and \eqref{eq_stepestDescent} is known as the steepest decent approach. Now, instead of using a first-order partial derivative, a \textit{second-order partial derivative} ($\nabla^2$) of cost function $c_f$ can be used as: 
\begin{equation}
\label{eq_Hessian}
H^t =  \nabla^2 c_f = \frac{\partial^2 c_f}{\partial \text{\bf{w}}},
\end{equation}  
where $H^t$ is \textit{Hessian} matrix at the $ t $-th iteration~\cite{annQnChen1994}. Hence, the weight change $\Delta \text{\bf{w}}^t$ using second--order Taylor's series expansion of cost function $c_f$ around point $\text{\bf{w}}^t$ is computed as: 
\begin{equation}
\label{eq_newtonUpdate}
\Delta \text{\bf{w}}^t = - H^{^t-1} g^t,
\end{equation}
where $ H^{^t-1} $ is the inverse of Hessian matrix $ H^t $ and the weight change $\Delta \text{\bf{w}}^t$ is known as the \emph{Newton method} or Newton update~\cite{haykin2009neural}. In the past, several algorithms were proposed using~\eqref{eq_gradient} and \eqref{eq_newtonUpdate}. Some of them are summarized as follows:

Backpropagation (BP) is a \textbf{first-order gradient-descent} algorithm for the FNNs optimization~\cite{annBpWerbos1975,annBpRumelhart1986}. In BP, the error computed at the output layer is propagated backward to the hidden layers. BP algorithm has two phases of computation: \textit{forward computation} and \textit{backward computation}, where at $ t $-th iteration, the weight change $ \Delta \text{\bf{w}}^t $ for $l$-th layer is computed as:
\begin{equation}
\label{eq_bpWC}
\Delta \text{\bf{w}}^t_{l} = \alpha^t\text{\bf{w}}^{t-1}_l + \eta^t g^t \text{\bf{y}}_{l-1},
\end{equation} 
where $\text{\bf{y}} $ is inputs/excitation from previous layer $ l - 1  $, $\eta^t$ is learning rate and  $\alpha^t$ is momentum factor. 

The choice of learning rate $\eta^t$ and momentum factor $\alpha^t$ are critical to gradient-descent technique. The momentum factor $\alpha^t$ allows BP training to be biased with previous iteration weights that help convergence rate to be faster. BP is sensitive to these parameters~\cite{annBpRumelhart1986}. If the learning rate is too small, learning will become slow, and if the learning rate is too large, learning will be zigzag and algorithm may not converge to required degree of satisfaction. Additionally, a high momentum factor leads to a high risk of overshooting minima and a low momentum factor may avoid local minima, but learning will be slow.  The classical BP algorithm is slow and has a tendency to fall in local minima~\cite{annMinimaGori1992problem}. 

Since the basic version of BP is sensitivity towards learning rate and momentum factor~\cite{annBpRumelhart1986}, several improvements were suggested by researchers: 1) a fast BP algorithm, called \textit{Quickpro} was proposed in~\cite{annBpFahlman1989cascade,annBpFahlman1988empirical}; 2) a \textit{delta-bar} technique and an \textit{acceleration} technique was suggested for tuning BP learning rate $\eta$ in~\cite{annBpJacobs1988increased} and in~\cite{annBpSilva1990acceleration} respectively; and 3) a variant of BP, called resilient propagator (\textit{Rprop}) was proposed in~\cite{annBpRiedmiller1993direct}. 

In the \textit{Rprop}, if the gradient direction in iteration $n$ remains unchanged from its previous iteration $t-1$, then the weight change will occur in larger magnitude, else in smaller. In simple words, if gradient sign remains unchanged from previous iterations, the magnitude of learning rate $ \eta $ will be large, otherwise small. The proposed \textit{Rprop} improves determinism of convergence to global minima~\cite{annBpRiedmiller1993direct}. However, it is not faster than the \textit{Quickpro}, but still faster than BP~\cite{annBpSchiffmann1994optimization}.

Contrary to BP, a \textbf{second-order minimization method}, called \textit{conjugate gradient}  (CG) can be used for weights optimization~\cite{annCgHestenes1952methods,annCGbarnard1989neural,annCgCharalambous1992}. The CG does not proceed down with a gradient; instead, it moves in the direction that is conjugate to the direction of the previous step. In other words, the gradient corresponding to the current step stays perpendicular to the direction of all the previous steps, and each step is at least as good as its previous step. Such series of steps are non-interfering. Hence, the minimization performed in one step will not be undone by any further steps. Several variants of the CG were proposed in the past~\cite{annCgDai1999nonlinear}.

Similar to the CG, many other variants of derivative-based conventional methods are used for weights optimization: \textit{Quasi-Newton}~\cite{annQnChen1994}, \textit{Gauss-Newton}~\cite{bertsekas1999nonlinear}, or \textit{Levenberg-Marquardt}~\cite{annMarquardt1963}. Quasi-Newton uses a second-order partial derivative~\eqref{eq_Hessian} of error~\eqref{eq_mse}, and it computes its weight search direction by using Broyden-Fletcher-Goldfarb-Shanno (BFGS) method~\cite{annCgFletcher1987practical}. In Gauss-Newton method, the FNNs optimization is framed as a nonlinear least square optimization problem, which suggests to using the sum of squared error~\eqref{eq_mse}~\cite{annMarquardt1963}. Many researchers suggested that the Levenberg-Marquardt (LM) method outperforms BP, CG, and Quasi-Newton methods~\cite{annLmHagan1994,annLmLera2002}. Several other methods were proposed for the FNNs optimization are based on \textit{Kalman-filter}~\cite{haykin2001kalman,sum1999kalman} and \textit{recursive least squares method}~\cite{azimi1992fast}. 

\subsection{Comments on conventional approaches}
The gradient-descent based conventional algorithms operate on a single solution (a weight vector) during the optimization. Thus, these algorithms are computationally faster than the algorithms that use two or more solution vectors during the optimization and select the best solution vector at the end of optimization iterations. Moreover, the gradient-decent methods such as BP~\cite{annBpRumelhart1986}, Online BFGS~\cite{schraudolph2007stochastic} can be applied for the stochastic as well as batch mode training of the FNNs. 

The basic advantages of the stochastic/online training of an FNN are its ability to address redundancy in training pattern, the inclusion of training data that are currently not in training set (i.e., the possibility of dynamic learning), and faster training than that of batch mode~\cite{wilson2003general}. Whereas, most of the other second-order gradient-descent methods and metaheuristic algorithms can only use batch mode training. However, a batch mode (offline) training of an FNN can at least guarantee a local minima under a simple condition compared to a stochastic/online training, and for a larger dataset, batch mode training can be faster than stochastic training~\cite{nakama2009theoretical}. 

On the contrary to the advantages of the conventional methods, they have several limitations. For example, they have the tendency to fall in local minima, and they are only used for optimizing the FNN weights. Primarily, the gradient-descent algorithms depend on the error function, e.g., mean square error or sum of squared error. For instance, the least square methods like the Gauss-Newton and LM works only if the cost function is the sum of squared error. The Newton method has to compute a Hessian matrix~\eqref{eq_Hessian}, which has to be positive definite and computing the Hessian matrix~\eqref{eq_Hessian} can be hard and expensive. Similarly, the Quasi-Newton and CG methods need to use a line-search method that sometimes can be expensive. 

Moreover, the FNN generalization, as mentioned in Section~\ref{sec_FNN_gen}, needs the reduction in the number of weights (less complex network architecture). Hence, the application of conventional algorithms is limited compared to the metaheuristic algorithms such as the genetic algorithm (GA) that can be directly applied to an FNN for its automatic structure determination and complexity reduction~\cite{gaHolland1975,gaGoldberg1988}. Similarly, a metaheuristic algorithm can formulate an FNN such that the insignificant weights of the network can be eliminated to improve the FNN generalization ability. Moreover, metaheuristic algorithms can evolve an FNN as a whole by optimizing its components simultaneously.  

\section{Metaheuristic approaches}
\label{sec_fnnMetaOpt}
\label{sub_noFreeLunch}
So far, only the gradient-descent based algorithms were discussed, which are local search algorithms. They are good at exploiting the obtained solutions to find  new solutions. However, to find a global optimum solution, any optimization algorithm must use two techniques: 1) \textit{exploration}---to search new and unknown areas in a search space, and 2) \textit{exploitation}---to take advantage of the already discovered solution~\cite{mhMarch1991exploration}. The exploration and exploitation are two contradictory strategies and a good search algorithm must find a trade-off between these two. Metaheuristic is the procedure that implements nature-inspired heuristics to combine these two strategies~\cite{mhOsman1996metaheuristics}. Hence, metaheuristic approaches are alternative to the conventional approaches for optimizing the FNNs. 

Unlike the conventional methods, which require the objective function to be continuous and differential, the metaheuristic algorithms have the ability to address complex, nonlinear, and non-differentiable problems. However, the optimization algorithms are often biased towards a specific class of problems, that is, ``there is no such universal optimizer which may solve all class of problem," which is evident from no free lunch theorem~\cite{mhWolpertMacready97nofree}.

Wolpert and Macready~\cite{mhWolpertMacready97nofree} introduced \textit{no free lunch} (NFL) theorem to answer the question, ``whether a general purpose optimization algorithm exists." Moreover, Wolpert~\cite{mhWolpert1996lack} introduced NFL for optimization algorithm to answer the question, ``How does the set of problems $F_1 \subset \mathcal{F}$ for which algorithm $a_1$ performs better than algorithm $a_2$ compares to the set $F_2 \subset \mathcal{F}$ for which the reverse is true." Here, $ \mathcal{F} $ is space of all possible problems. To answer this question, Wolpert proposed NFL theorem, which says that ``the average performance of any pair of algorithms across all possible problems is identical"~\cite{mhWolpertMacready97nofree}. 

Therefore, a straightforward interpretation of NFL is as follows. ``A general purpose universal optimization strategy is impossible, and the only way one strategy can outperform another if it is specialized to the structure of the specific problem under consideration"~\cite{nhHo2002simple}. Schumacher et al.~\cite{schumacher2001no} argue that the NFL theorem~\cite{mhWolpertMacready97nofree} holds only for the set of problems which are closed under permutation (c.u.p). Therefore, indeed the performance of one algorithm can be improved over another for the problems that are not c.u.p and most of the real-world problems are not c.u.p~\cite{igel2003onclasses}. Such is the reason why in the past researchers were inclined to create and improvise algorithms for optimizing the FNNs.

\subsection{Metaheuristic algorithms}
Since a large  variety of metaheuristic algorithms are available, it is difficult to classify  metaheuristic algorithms precisely into different classes. Though, intuitively, three basic categorize can be done:
\subsubsection{Single solution based algorithms }
A single solution based metaheuristic algorithm operates on a single solution (candidate) and applies some heuristic inspired by the nature or some other phenomena on the current solution. For example, algorithms like simulated annealing (SA)~\cite{saKirkpatrick1983}, tabu search (TS)~\cite{tsGlover1989tabu}, variable neighborhood search (VNS)~\cite{vnsMladenovic1997variable}, greedy randomized adaptive search (GRAP)~\cite{mhFeo1995greedy}, etc., improves a single solution by searching around its neighborhood and continue to improve the solution until a satisfactory solution is obtained. For instance, the heuristics of some algorithms are as follows:

SA is a probabilistic approach that imitates the cooling strategy (annealing process) of a metallurgy industry. It uses Monte Carlo method~\cite{saMetropolis1953} to determine the acceptance probability of a newly generated solution in the neighborhood of the current solution. Hence, for a given search space, SA should guide a solution towards a global optimal solution~\cite{saKirkpatrick1983,saCerny1985thermodynamical}. Similarly, TS  is inspired by the human behavior of tabooing objects~\cite{tsGlover1989tabu}. In other words, TS discourages (tabu) the acceptance of the solutions that are already explored in the past. Hence, it improves upon SA by introducing some additional restriction on the acceptance of the new solutions. 

Since a single solution based algorithm exploits the current solution, it also is known as the \textit{local search algorithm}. The focus of this article is to illustrate the application of the metaheuristic algorithms for the FNNs optimization. Hence, for the detailed description of the mentioned algorithms, the readers may explore the respective references.

\subsubsection{Population based algorithms}
Population based algorithms operate on the multiple solutions (candidates) and apply the heuristics inspired by nature, biological evolution, biology, or some other forms. In contrast to the single solution based algorithms, the population based algorithms have a high exploration (global search) ability. The following are the population based algorithms:  

\paragraph{Evolutionary algorithms (EA)} Genetic algorithms (GA)~\cite{gaHolland1975,gaGoldberg1988}, evolutionary programming (EP)~\cite{eannFogel1998evolutionary}, evolutionary strategy (ES)~\cite{schwefe1987evolutionstrategie}, genetic programming (GP)~\cite{eannKoza1999GP}, differential evolution~\cite{deStorn1997}, etc., are the algorithms inspired by the dynamics of natural selection and use the operators, such as \textit{selection}, \textit{crossover}, and \textit{mutation} to find a near-optimal solution~\cite{gaGoldberg1988}. EA framework offers an exploration of a vast search space and guarantees to find a near-optimal solution. Since EAs do not depend on gradient information, they solve a large range of complex, nonlinear, nondifferentiable, multimodal optimization problems. Also, EAs give a wider scope in the FNN optimization since EAs can optimize both discreet and continuous optimization problem, and the FNN components can be formulated into both ways.

The differences between EAs can be briefly stated as follows: GA uses genetic operators, such as selection, crossover, and mutation to search optimum genetic vector from a search space~\cite{gaGoldberg1988}; whereas, only the mutation operator are used in ES to evolve a real vector solution~\cite{schwefe1987evolutionstrategie}. On the other hand, GP searches an optimum program structure from a topological search space of computer programs~\cite{eannKoza1999GP} and EP are used for evolving parameters of a computer program whose structure is kept fixed~\cite{eannFogel1998evolutionary}. 

\paragraph{Swarm intelligence (SI)} SI algorithms are inspired by the collective and self-organized behavior of the swarm (insects, birds, fish, etc.). Particle swarm algorithm (PSO)~\cite{psoEberhart1995}, ant colony optimization (ACO)~\cite{acoDorigoColorni1996}, artificial bee colony (ABC)~\cite{abcKaraboga2005}, bacterial foraging optimization (BFO)~\cite{bfPassino2002biomimicry}, etc., are some widely used SI algorithms. The basic principle of SI algorithms is as follows. First, a swarm (collection of solutions) are randomly generated. Then, the heuristic inspired by the swarm behavior modifies the current solution. For example, in PSO, ACO, ABC, and BFO, the heuristics are inspired by the foraging behavior of birds, ants, bees, and bacteria respectively. In these algorithms, an FNN component is formulated as a solution for the optimization. 

In PSO, a swarm, as a whole, is like a flock of birds (particles, which are the weight vectors) that collectively foraging (explore search space) for food (best weight vector) and is likely to move close to an optimum food-source~\cite{psoEberhart1995,psoKennedy2001swarm}. Moreover, each particle bears two properties: location and velocity. The location of a particle is a solution vector (weight vector $ \text{\bf{w}} $), and velocity  $\eta$ is a vector equal to the size of the location vector. Each particle determines its movement using knowledge of its best locations, global best location, and random perturbations~\cite{psoEberhart1995,psoShi1998modified}. 

In ACO, the artificial ants explore the area around their nest (colony) for searching a food source. ACO takes advantage of ants ability to choose the shortest path to a food source by communicating among each other's using a pheromone secretion~\cite{acoDeneubourg1990}. This behavior of ants led to the development of ACO algorithm~\cite{acoDorigoColorni1996}. 

Similarly, in ABC, three kinds of honey bees, such as employed bee, onlooker bee, and scout bee are responsible for searching food source~\cite{abcKaraboga2005}. Each employed bee memorizes a food source, i.e., a solution (weight vector). Then, each onlooker bee examines the nectar amount (fitness of solution) of a food source memorized by the employed bees and depending on nectar amount; they send scout bees for searching new food source. Hence, they iteratively construct the solution. The readers are encouraged to explore the detail description the algorithms in their respective references.

\paragraph{Other metaheuristics} The population based metaheuristic algorithms metaphor is exploited to device several algorithms. There are algorithms inspired by the behavior of animals, birds, and insects, such as gray wolf optimization (GWO)~\cite{mhMirjalili2014grey}, flower pollination (FP)~\cite{mhYang2012flower}, cuckoo search (CS)~\cite{csYang2009cuckoo}, firefly (FF)~\cite{ffYang2010nature}, etc. 

Similarly, there are algorithms inspired by some phenomenon observed in the physics and chemistry, such as harmony search (HS)~\cite{hsZongWooGeem2001}, central force optimization (CFO)~\cite{cfoFormato2007}, gravitational search optimization~(GSO)\cite{gsRashedi2009}, etc. 
A detailed list and classification of metaheuristic algorithms are provided in~\cite{mhFister2013brief}. 

The growing number of metaheuristic algorithms has drawn researchers to examine the metaphor, the novelty, and the significant differences among the metaheuristic algorithms~\cite{weyland2010a,sorensen2015metaheuristics}. In~\cite{sorensen2015metaheuristics}, the author provided an insight of the metaheuristic developed over the time, starting from SA to TS, EA, ACO, HS, etc. The author claimed that most of the metaheuristic are similar in nature and do not provide a groundbreaking method in optimization. Despite the criticisms, the author acknowledged the quality of metaheuristic research has been produced and can be produced. 

\subsubsection{Hybrid and memetic algorithms}
\label{subsec_hybrid}
An effective combination of various metaheuristic algorithms may offer a better solution than that of a single algorithm. A paradigm of hybrid algorithms, called \textit{memetic algorithm} gave a methodological concept for combining two or more metaheuristics (global and local) to explore a search space efficiently and to find a global optimal solution~\cite{mhMoscato1989evolution}. 

\begin{figure}
	\centering
	\includegraphics[scale=0.8]{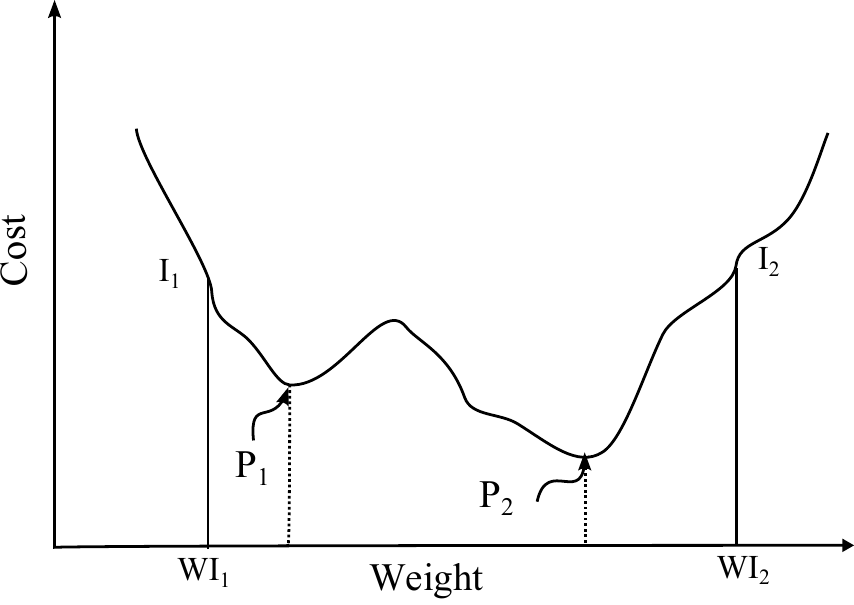}
	\caption{Metaheuristic may be used for finding initial weights $ \text{WI}_{2} $ and conventional algorithms may be for finding global optima $ \text{P}_{2} $ and vice versa~\cite{eannXinYao1993}.}
	\label{fig_hybrid}
\end{figure}

The conventional algorithms have the local minima problem because they lack global search ability, but they are fast and good in local search. On the other hand, the metaheuristics are good in global search, but they suffer premature convergence\cite{leung1997degree,trelea2003particle}. Therefore, a combination of these two may offer a better solution in FNN optimization than that of using any one of them alone (Fig.~\ref{fig_hybrid}). To reach a global optimum, a hybrid strategy can be used. Fig.~\ref{fig_hybrid} shows an impact of hybridization of metaheuristics on the FNNs optimization. The hybrid algorithms can be categorized in two paradigms:
\begin{enumerate}[1)]
	\item The combination of conventional and metaheuristic algorithms---to take advantage of local search and global search algorithms.
	\item The combination of two or more metaheuristic algorithms---to make use of different heuristics or a combined influence of two or more heuristics of global search algorithms.
\end{enumerate}

Under the definition of memetic algorithms, researchers combine EAs with conventional algorithms~\cite{brownlee2011clever}. For example, the effectiveness of global (GA) and local search (BP) combination is explained in~\cite{yin2011hybrid,yaghini2013hybrid}. Similarly, a hybrid PSO and BP algorithms for optimizing FNN were found useful in obtaining better approximation than using one of them alone~\cite{psoZhang2007}. 

A convergence scenario similar to Fig.~\ref{fig_hybrid} was illustrated in~\cite{abcOzturk2011}, where ABC was applied for searching initial weights and LM was applied for optimizing the already discovered weights. An example of effectively combining two metaheuristic GA and PSO is illustrated in~\cite{psoJuang2004}, where both GA and PSO optimized the same population. A detailed description of the hybrid metaheuristic algorithms for the FNN optimization is described in the following Section.
  
\subsection{Metaheuristic formulation of the FNN components}
\label{sub_formulatingFNN}
Metaheuristics are stochastic/non-deterministic algorithms. Hence, they do not guarantee a global optimal solution, but they can offer a near-optimal (satisfactory) solution. Moreover, metaheuristics efficiently solve a wide range of complex continuous optimization problems; especially when the problems have incomplete and imprecise information. 

The basic form of FNN optimization is the act of searching its weights (free parameters of FNN) such that the cost function~\eqref{eq_mse} or similar function can be minimized. However, the goodness (performance) of FNN cost function depends not only on finding optimum weights, but finding the optimum architecture, activation function, parameter setting of learning algorithm, and training environment are equally important. To apply metaheuristic algorithms for optimizing an FNN, its components (\emph{phenotype}) need to be formulated into a vector (\emph{genotype}) form.

\begin{figure}
	\centering
	\includegraphics[scale=0.75]{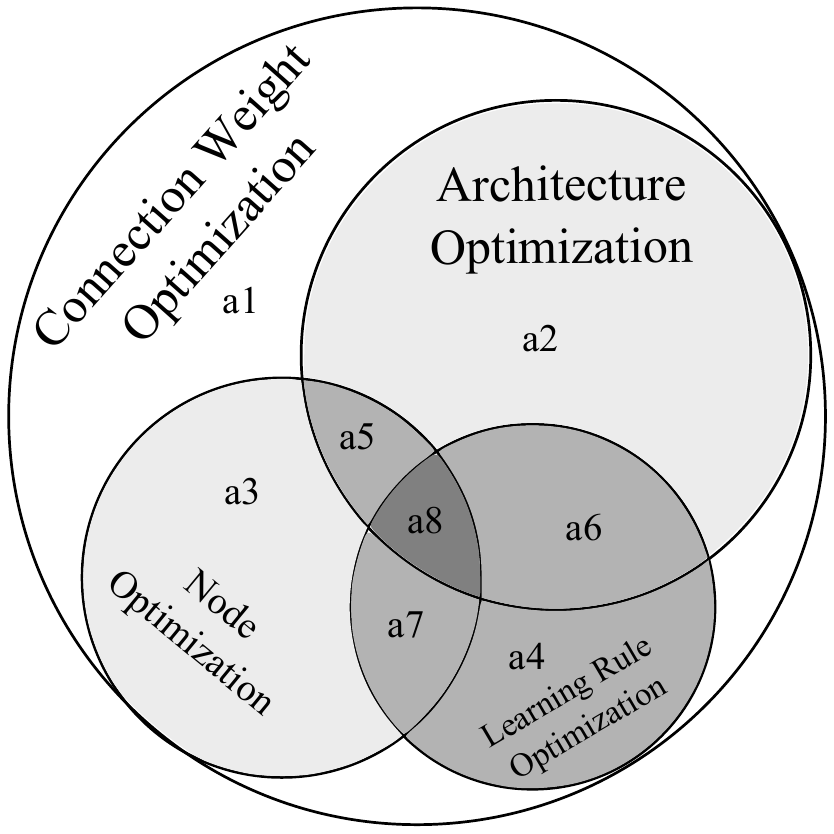}
	\caption{Spectrum of metaheuristic design of FNN}
	\label{fig_mhFNN}
\end{figure}

Usually, the FNN components, such as weights, architecture, activation function, learning rule, etc., are considered arbitrarily. Then, a learning algorithm is applied to search weights while the other components are kept fixed to their initial setting. The metaheuristics, on the other hand, allow us to optimization each component simultaneously or a combination of components efficiently (Fig.~\ref{fig_mhFNN}). 

The \emph{Venn diagram} in Fig.~\ref{fig_mhFNN} illustrates the spectrum of FNN components optimization: weights, architecture, activation function, and learning rule's parameters. In Fig.~\ref{fig_mhFNN}, area ``a1" indicates the optimization of weights; area ``a2" indicates the optimization of weights and architecture; area ``a3" indicates the optimization of weights, architecture, and activation function; and all other possible combinations. Examining Fig.~\ref{fig_mhFNN}, one can say that the strength and complexity of optimization increases from area denoted ``a1" to ``a8," where ``a1" is the simplest approach and ``a8" is the most sophisticated approach.  

Each FNN component can be separately optimized on a one-by-one basis. Therefore, firstly, the weights can be optimized by keeping a fixed architecture; secondly, the architecture can be optimized keeping weights fixed; thirdly, the activation function can be optimized keeping architecture and/or weights fixed; and so on. Another way is to optimize all or a combination of FNN components simultaneously. Therefore, weights and architecture can be optimized, simultaneously; or weights and activation functions, simultaneously; or weights, architecture, and activation functions simultaneously; and so on. In the simultaneous optimization of all or a combination of components, a vectored representation of all components, or a combination of components can be optimized respectively. Once a vectored representation (genotype) is designed, then one of the available metaheuristic algorithms in the literature can be applied to optimize the designed vector to obtained an optimum FNN.  

Now, there are \emph{single-solution} based and \emph{population} based metaheuristics~\cite{mhSurveyBoussaid2013}. In a single-solution based metaheuristic algorithm, a genotype $ \text{\bf{w}} = \langle w_1, w_2, \ldots, w_n \rangle $ is used. Whereas, in a population-based metaheuristic algorithm, a collection of many genotypes are used. In other words, a population matrix $\text{\bf{W}} = (\text{\bf{w}}_1, \text{\bf{w}}_2, \ldots, \text{\bf{w}}_m)$ of $m$ weight vectors is used. 

Yao~and~Liu~\cite{eannYao1998} identified evolution at various components of FNN that fall into the spectrum of metaheuristic design of FNN shown in Fig.~\ref{fig_mhFNN}. This Section will describe \textit{how researchers  applied metaheuristics for evolving FNN}. The evolution in FNN components is described here one-by-one, as follows. Here, the word optimization, adaptation, and evolution are used in the similar context.

\subsubsection{Weight optimization}
\begin{figure}
	\centering
	\subfigure[]{
		\label{fig_FNN}
		{\includegraphics[scale=0.7, angle = 0]{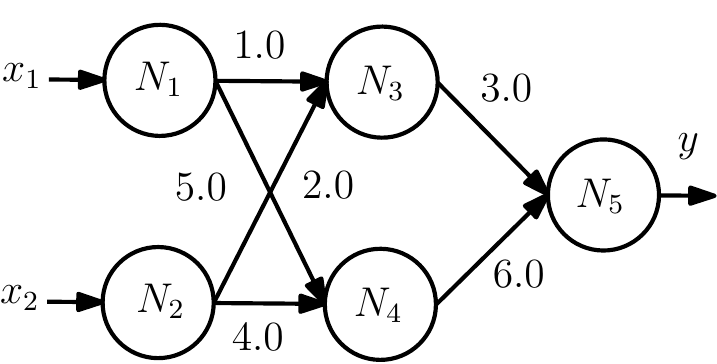}}
	}~
	\subfigure[]{
		\includegraphics[scale=0.8]{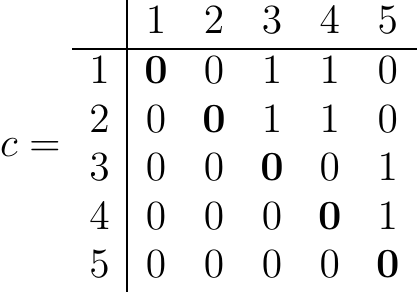}
		\label{fig_ConFNN}
	}~
	\subfigure[]{
		\includegraphics[scale=0.8]{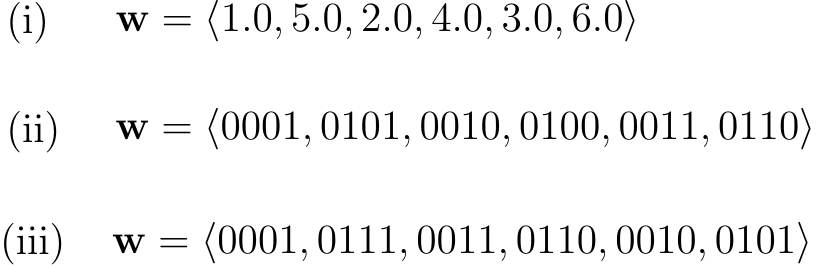}
		\label{fig_GenoConn}
	}		
	\caption{Mapping of phenotype to genotype. (a) Phenotype of a three-layer FNN. (b) Adjacency matrix. (c) Weights encoding: i) real value; ii) 4-bits binary; iii) 4-bits gray encoding.}
	\label{fig_pheno_to_geno}
\end{figure}

FNN weight optimization is the most common and widely studied approach, in which the weights are mapped onto an $ n $-dimensional weight vector $ \text{\bf{w}} $, where $ n $ is the total number of weights in a network. The vector $ \text{\bf{w}} $ is a genotype representation of a phenotype (FNN structure), where the weight $ \text{\bf{w}} \in \mathbb{R}^n $. The weights $ w_i $, an element of vector $  \text{\bf{w}}$, may be encoded in the following ways: by assigning a real value, $ l $-bits binary coding, $ l $-bits gray coding, IEEE floating point coding, etc. Fig.~\ref{fig_pheno_to_geno} is an example of phenotype to genotype mapping, where a phenotype shown in Fig.~\ref{fig_FNN} that has the connectivity matrix $ c $ as per Fig.~\ref{fig_ConFNN} is encoded into three different weight vectors shown in Fig.~\ref{fig_GenoConn}.  

FNN weights optimization using metaheuristic is practiced from \textbf{early 80's }when even the term metaheuristic was not used. Engle's~\cite{saJonathanEngel1988} work on FNN weight vector optimization using SA was the first evidence of metaheuristic application. To optimize weight vector using SA, first, the phenotype was mapped onto a real-valued weight vector (Fig.~\ref{fig_GenoConn}), and to compute fitness of the FNN, a \emph{reverse mapping} from genotype (weight vector $ \text{\bf{w}} $) to phenotype (FNN) was used. Such process was continued until a satisfactory solution was found. SA based FNN weight optimization was found to be performing better in comparison to conventional approaches~\cite{saShang1996,saSexton1999,saSarkar2003}. 

Similar to Engle's~\cite{saJonathanEngel1988} approach of phenotype to genotype mapping and vice versa, in~\cite{tsBeyer1991}, the FNN weights optimization was performed using TS. In~\cite{tsBattiti1995}, an improvised TS, called reactive tabu search was used for optimizing weights. Several studies show that TS when used for optimizing FNN weights, outperformed BP and SA algorithm~\cite{tsSexton1998global,tsYe2007}. However, SA and TS are single solution based algorithms, which has a limited scope of exploring search space to obtain a global optimal solution. In contrast, the EAs, SI, or other bio-inspired metaheuristics are population-based algorithms that operate on multiple agents to explore a search space. Hence, they have a better exploration ability than SA, TS, BP, CG, and other single solution based algorithms~\cite{gaGoldberg1988,psoKennedy2001swarm}.

For optimizing the FNN weights, EAs use two \textbf{types of vector representation}: real-valued and binary valued vector representation. In Fig.~\ref{fig_GenoConn}, following weight vector representation are illustrated: 1) real-valued coded chromosome, 2) binary coded chromosome, and 3) binary gray-coded chromosome. 

Goldberg and Holland~\cite{gaGoldberg1988} gave the idea of FNN training using GA. However, Whitley and Hanson~\cite{gaWhitley1989theGENITOR} were the first to propose ``GENITOR,'' a GA based FNN training procedure that used binary-coded chromosome (Fig.~\ref{fig_GenoConn}) for optimizing the weights. Many others followed the idea of GENITOR with some additional improvements such as connectivity optimization and reduced search space introduction~\cite{gaWhitley1990,eannSrinivas1991learning}. On the other hand, in~\cite{eannBelew1990evolvingnetworks}, a binary gray coding (Fig.~\ref{fig_GenoConn}) scheme was used for optimizing the weights, where at first, GA was used for finding initial weights that were further optimized by using BP and vice versa. 

The binary bit-string representation of the weights leads to a \textbf{precision problem}, i.e., \textit{how many bits would be sufficient for representing weights} and \textit{what would be the total length of a chromosome}. Moreover, the binary representation is computationally expensive because, in each training iteration, a binary to real-valued mapping and vice versa is required. Hence, it is advantageous to use the real-coded chromosome (Fig.~\ref{fig_GenoConn}) directly~\cite{gaMontana1989,eannFogel1990evolving,eannSietsma1991creating,eannMenczer1992evidence,LGARirani2011evolving}. 

Traditional \textbf{EA operators} are applied on the binary chromosome. Thus, operators, such as bias-mutation, unbiased-mutation, node-mutation, weight-crossover, and gradient-operator, etc., were defined, for operating on the real-valued chromosome~\cite{gaMontana1989}. On the other hand, a matrix-based representation of weights, where a column-wise and a row-wise crossover operators were also defined~\cite{gaDongsun2005}. The GA-based real coded weights optimization outperforms BP and its variants for solving real-world applications~\cite{gaKitano1990,gaSexton1998,LGA1ding2011optimizing,LGA2tong2010genetic}. Moreover, an evolutionary inspired algorithm, called differential evolution (DE)~\cite{deStorn1997,deDas2011differential} that imitates mutation and crossover operator to solve complex continuous optimization problems was found to be performing efficiently for real-valued weight vector optimization~\cite{eannNolfi1994learning,deIlonen2003differential,deSlowik2011application}.

Similar to DE, \textbf{swarm-based or bio-inspired based metaheuristics} directly apply heuristics inspired by nature on a real-valued vector. Hence, they are advantageous in comparison to an EA-based algorithm that needs to simulated mutation and crossover operators for real-valued weight vector~\cite{psoKennedy2001swarm}. It was found that \textbf{PSO} guides a population of  the FNN weight vectors towards an optimum population~\cite{psoIsmail2000,zhang2007hybrid}. Hence, many researchers resorted to working on swarm based metaheuristics for the FNN optimization. 

A \emph{cooperative PSO}, which suggests to splitting a solution vector into $ n $ parts, where each part optimized by a swarm of $ m $ particles~\cite{psoBergh2001,van2004cooperative}. Thus, an $ n \times m $ combinations are constructs an $ n $-dimensional composite vector, where each $ m $ swarm contributes to the fitness of a solution. Such cooperation between swarms led to a better performance than that of the basic version of PSO. Similarly, a \emph{multi-phase PSO} was proposed in which particle position was updated only when improvement in location was found; otherwise, the location was copied as-it-is into the next generation~\cite{psoAl-kazemi2002}. 

A \emph{cultural cooperative particle swarm optimization} (CCPSO) approach in which a collection of multiple swarms that interact by exchanging information was proposed by Lin et al.~\cite{psoChengJian2009}. The CCPSO performed better than BP and GA when it was applied for optimizing a fuzzy neural network. Similarly, a hierarchical particle swarm optimization was used to design a beta basis function neural network~\cite{psoAjithPSO2013}. 

Apart from PSO, there are numerous metaheuristic algorithms among which, some significant metaheuristics were discussed here that were applied for FNN optimization. The continuous version of \textbf{ACO}~\cite{acoSocha2008} was efficiently applied to optimize the FNN weight vector~\cite{acoKrzysztofSocha2007}. ACO trained FNN was found efficient in solving real-life applications, such as scheduling, prediction, image recognition, etc.~\cite{acoIrani2012,acoSharma2013ant}. 

\textbf{ABC} was efficiently applied on weight vector for optimizing the FNNs~\cite{abcKaraboga2007,abcGarro2011trn,abcSarangi2014trn}. Similarly, 
considering the efficiency of \textbf{HS} algorithm---that has a slow convergence rate, but guarantees a near-optimum solution~\cite{hsZongWooGeem2001}---many researchers applied HS for optimizing weight vector of the FNNs~\cite{hsKattan2010,hsKulluk2012}. Moreover, the efficiency of HS comes from using $m$ many harmonies (weight vectors), and iteratively improvising each harmony by computing new harmony (new solution vectors) using heuristic inspired by music pitch modification~\cite{hsZongWooGeem2001,hsMahdavi2007,hsPan2010}.

In the past, many \textbf{other forms of metaheuristics} were also used for optimizing the FNNs. For example, the application of FF, CS, GSO, BFO, and CFO algorithms for the FNN weights optimization is available in~\cite{ffHorng2012firefly}, \cite{csVazquez2011snn}, \cite{gsGhalambaz2011hybrid}, \cite{bfUlagammai2007,bfZhang2010bacterial}, \cite{psoGreenII2012} respectively. 

Moreover, a comparative study showed that FF algorithm performed better than that of BP, GA, and ABC for weight vector optimization~\cite{ffNandy2012training}. In~\cite{mhAlba2006metaheuristic} a detailed study explains the application of the local and global metaheuristic algorithm for FNN optimization. For example, local search algorithms like SA, TS, GRAP, VNS~\cite{vnsMladenovic1997variable}, estimations of distribution algorithm~\cite{edaLarranaga2002} and global search algorithms like GA, ACO, and memetic algorithm, were examined thoroughly in~\cite{mhAlba2006metaheuristic}. Additionally, many researchers studied the performance of metaheuristic algorithms for the training of the FNN and reported that the metaheuristic approaches outperform all the conventional methods by a huge margin~\cite{carvalho2011metaheuristics,kordik2010meta,khan2012comparison}. 

The \textbf{memetic algorithm} supports the hybridization of two or more global metaheuristics for the FNN optimization, which is  evident from the following examples. A hybrid GA and PSO approach for optimizing the FNN were proposed in~\cite{psoJuang2004}, where both GA and PSO were suggested to run over the same population---randomly generated population $ \text{W} $ of $m$ individuals (the same individual was treated as a chromosome in GA and a particle in PSO). In each generation of GA and PSO, the fitness of each individual was computed. Then, the best performing individuals (top-half) were marked as elites. The elite individuals were copied to next generation and half of the copied elites were optimized using PSO and the remaining half using GA through tournament selection and crossover operation. 

Similarly, a PSO and SA based \textbf{hybrid algorithm} for optimizing FNN were proposed in~\cite{psoDa2005}, where, in each iteration, each PSO particle was governed by SA metropolis criteria~\cite{saMetropolis1953} that determined global best particle for PSO algorithm. There are several other hybrid algorithm examples available in the literature: a hybrid PSO and GA~\cite{ali2013reservoir}; hybrid GA and DE~\cite{donate2013time}; hybrid PSO and GSO~\cite{mirjalili2012training}; and hybrid PSO and optimal foraging theory~\cite{psoNiu2007}. 

\subsubsection{Architecture plus weight optimization}
\begin{figure}
	\centering
	\subfigure[]{
		\label{fig_ArcDirect}
		{\includegraphics[scale=0.7, angle = 0]{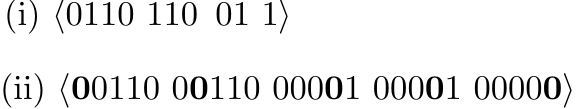}}
	}~
	\subfigure[]{
		\includegraphics[scale=0.75]{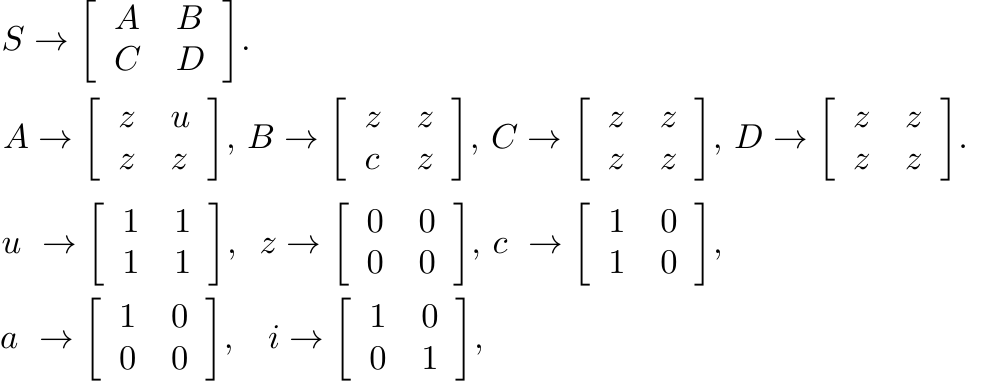}
		\label{fig_ArcIndirect}
	}~
	\subfigure[]{
		\includegraphics[scale=0.75]{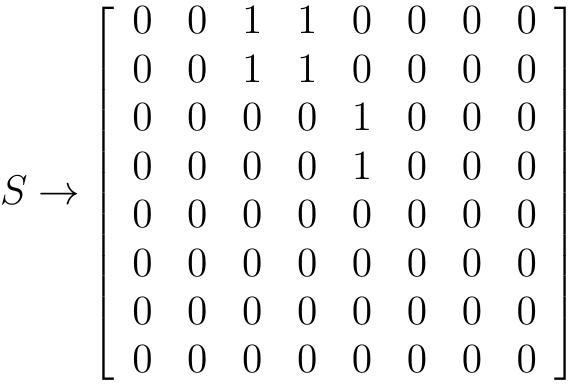}
		\label{fig_ArcIndirect1}
	}		
	\caption{Mapping of phenotype (Fig.~\ref{fig_pheno_to_geno} to genotype (for architecture). (a) Direct encoding to a vector of connectivity matrix (Fig.~\ref{fig_ConFNN}): i)  upper right triangle; ii) complete connectivity. (b) Indirect encoding schemes for architecture (Fig.~\ref{fig_FNN}), where $S$ is a start symbol, $A, B, C,$ and $D$ are the variables, and $a, c, i,$ and $u$ is the terminal. (c) Complete connectivity derived from rules operation shown in Fig.~\ref{fig_ArcIndirect}.}
	\label{fig_pheno_to_geno_Arch}
\end{figure}

The basic architecture optimization approach is a \textit{cascade correlation learning}, which iteratively adds nodes to hidden layer to construct optimum architecture~\cite{annBpFahlman1989cascade}. Moreover, a \textit{constructive} (add node iteratively) and \textit{destructive} (prune nodes iteratively) method~\cite{eannFrean1990upstart}. However, the constructive and the destructive methods for optimizing architecture are no different from the manual \textit{trial-and-error} method. Therefore, genetic representation of the FNN architecture as mentioned in Figs.~\ref{fig_ArcDirect}, \ref{fig_ArcIndirect}, \ref{fig_ArcIndirect1} can be used for architecture optimization, which is equivalent to searching optimum architecture from a compact space of FNN topology~\cite{eannManiezzo1994,LGA1xi2013architecture}. 

Let us discuss the genetic representation of architecture in detail. A \textbf{direct encoding scheme} (Fig.~\ref{fig_ArcDirect}) was proposed in~\cite{gaWhitley1989,eannSchaffer1990using}, where authors used an adjacency matrix (Fig.~\ref{fig_ConFNN}) to represent connections between nodes, where between any two nodes $ i $  and $ j $, a presence of connection is indicated by ``1", and absence of connection is indicated by ``0". Hence, they were able to encode complete structural information into a chromosome. However, it is disadvantageous because chromosome length increases with network size. Therefore, if only the network's structural information can be encoded into genotype, then, it will avoid chromosome length problem~\cite{eannHarp1989}. Additionally, the encoded network structural information can be accessed using rule-based recursive equation~\cite{eannMjolsness1989scaling}. Moreover, the represented parametric/structural information into the chromosome can indirectly provide access to the rest of the structural details from a predefined archive (parametric information)~\cite{eannKitano1990designing}. 

The \textbf{indirect encoding scheme} reduces chromosome length, where parametric information, such as the number of hidden layers, the number of nodes at hidden layers, the number of connection, etc., makes an archive $ s $. The production rule (Fig.~\ref{fig_ArcIndirect}) allow us to get access to complete structural information (Fig.~\ref{fig_ArcIndirect1}). Hence, a rule based encoding scheme allows a better FNN architecture optimization than a direct encoding scheme~\cite{eannSiddiqi1998comparison}. 

Unlike the weight optimization that has only limited ways of genetic representation, the FNN architecture optimization is an interesting area of research as there are various ways to represent architecture into genotype. It is evident from a fractal configured FNN representation in~\cite{eannMerrill1991fractally}, where authors define each node using parameters, namely, edge code, input coefficient, and output coefficient. Similarly, in~\cite{eannAndersen1993constructive}, GA was applied to evolve each layer separately and in~\cite{eannTayefehMahmoudi2013}, a grammar encoding and colonial competitive algorithm were proposed.

Another approach to the genetic representation of architecture is to encode weights $ \text{\bf{w}} $ (real vector: Fig.~\ref{fig_GenoConn}) and architecture vector $ \text{\bf{a}} = \langle a_1, \ldots a_m \rangle $ (binary vector as Fig.~\ref{fig_ArcDirect}) into a combined genotype. Hence, a single solution vector $ s = \langle w_1,\ldots,w_n, a_1,\dots,a_m \rangle $ is obtained~\cite{saLudermir2006}, which can be optimized by using metaheuristics. 

Many researchers improvised the algorithms itself to optimize architecture. Such examples are as follows: in~\cite{psoCarvalho2007}, a \textbf{PSO-PSO} method was proposed, in which a PSO (inner PSO block) was applied for optimizing weights that were nested under another PSO (outer PSO block) which was applied for optimizing the architecture of FNN by adding or deleting hidden node. Similarly, in~\cite{othrTsai2004hybrid,othrTsai2006training}, a hybrid Taguchi-genetic algorithm was proposed for optimizing the FNN architecture and weights, where authors used a genetic representation of the weights, but they select structure using constructive method (by adding hidden nodes one-by-one). A multidimensional PSO approach was proposed in~\cite{psoKiranyaz2009} for constructing FNN automatically by using an architectural (topological) space. Moreover, the individuals in the swarm population were designed in such a way that it optimized both position (weights) and dimension (architecture) of an individual in each iteration. Thus, optimized FNN weights and architecture simultaneously.

So far, only genetic representation was discussed for evolving architecture. However, \textbf{GP} can optimize a phenotype itself, where genetic representation is not required~\cite{eannMahsalKhan2013}. Therefore, EP and GP can be directly applied to a population FNN architecture to evolve an optimum FNN architecture~\cite{eannFogel1990evolving,gpKoza1991genetic,ensmblTsoulos2008}. 

The design of the FNN architecture is responsible for processing high-dimensional data. Hence, \textbf{deep learning} paradigm offers study massive and deep structure of the neural network that can process complex problems related to speech processing, natural language processing, signal processing, etc.,~\cite{schmidhuber2015deep,lecun2015deep}. Such a variant of the FNN is \textit{convolutional neural networks} (ConvNets), which is designed to process data from the multiple arrays form such as a color image composed of three 2D arrays~\cite{schmidhuber2015deep,lecun2015deep}. The ConvNets has a three-dimensional arrangement of neural nodes. Hence, it efficiently receives 3D inputs and processes them to produce 3D outputs~\cite{maturana2015voxnet}. 

In contrast to deep network paradigm, an \textbf{extreme learning machine} (ELM) based hierarchical learning framework (H-ELM) proposed in~\cite{tang2016extreme} claimed a faster learning than deep learning by ELM~\cite{huang2014insight} based auto-encoding. The proposed H-ELM framework worked in two phases: unsupervised hierarchical feature
representation and 2) supervised feature classification~\cite{tang2016extreme}.       

\subsubsection{Input layer optimization}
Input layer optimization resembles feature reduction, which is traditionally performed separately by dimensionality reduction methods~\cite{fodor2002survey}. However, reducing input dimension by optimizing input layer, i.e., by feeding a subset of input features at the input layer than by feeding the whole set of input features enhances FNN's performance~\cite{eannFontanari1991evolving,eannGuo1992using}. Therefore, FNN has a functional dependency on the problem at hand. 

EAs select a subset of input features for which FNN perform better than that of the complete feature set~\cite{eannFontanari1991evolving}. For this purpose, a genetic representation of input features is required in which the available features are placed on a genetic strip, and the presence of a feature is marked as ``1" and the absence of a feature is marked as ``0." Such mechanism of input layer optimization was found advantageous in improving NN performance~\cite{LGA1GAvenkadesh2013genetic}. 

Moreover, binary PSO~\cite{kennedy1997discrete}, which is a \textbf{discrete optimization method} was employed for selecting the input features which were binary coded~\cite{lin2008particle,vieira2013modified}. In~\cite{lin2008particle}, a modified version of binary PSO was proposed where EA like mutation operator was applied to mutated binary vectors. Similarly, ACO, which traditional solves discrete optimization problem was applied to select input features and training of an FNN in a hybrid manner~\cite{sivagaminathan2007hybrid}. 

Input layer optimization which is related to input feature reduction can also be thought as \textbf{training data optimization}. Training data optimization is helpful, particularly when data is insufficient or noisy. In~\cite{eannZhang1991neural}, an adaptive selection of input examples was performed by employing genetic selection, where two-point and one-point crossover operations created new example patterns. For the crossover operations, the parent's examples were drawn from the original input set. Also, mutation operators were also applied for generating new child example. Hence, the efficiency of FNN was improved when trained over the modified new examples. 

Additionally, in~\cite{eannCho1996evolution}, an \textbf{input example generation} methods was proposed in which the input space was divided into many regions, and $ k $-nearest neighbor method was applied to determine/generate a new virtual example, mainly for the sparse region of the input space. Hence, both the above methods of input example generation sought to enrich knowledge space for the FNN learning~\cite{eannZhang1991neural,eannCho1996evolution}.

\subsubsection{Node optimization}
Primarily, node optimization can be addressed in three ways: 1) by choosing activation functions at the FNN active nodes from a set of activation functions~\cite{eannLiu1996evolutionary,LGA2tong2010genetic}; 2) by optimizing the arguments of activation function~\cite{ojha2014simultaneous}; and 3) by placing a complete model at the nodes of a network~\cite{oh2002design,hirose2006complex}. 

It was found that FNN performed better when it has \textbf{non-homogeneous nodes} (different activation function at different nodes) than that of the homogeneous nodes~\cite{eannMani1990}. In~\cite{eannLiu1996evolutionary}, evolution in FNN nodes were offered by selecting sigmoid and Gaussian function adaptively at the nodes~\cite{eannLiu1996evolutionary}. Moreover, adaptation in both nodes and architecture using EAs, where the design of nodes was inspired by locus flight system and tailflip of crayfish~\cite{eannDumont1986neuronal}, can further improve FNN performance~\cite{eannStrok1990}. For this purpose, in~\cite{gaLing2007}, an input dependent FNN that had a combined chromosome representation (Fig.~\ref{fig_FNNgenEncoding}) was proposed, where a real-coded GA for simultaneous optimization of weights, activation functions, and architecture was used. 

On the other hand, to optimize nodes, a \textbf{family competitive EA} was proposed in~\cite{gaJinnMoonYang2001}, where three operators, such as decrease-Gaussian-mutation, Cauchy-mutation, and self-adaptive-mutation were defined. Moreover, family-competition is a process that generates a pool of $L$ many FNNs by recombination and mutation operations and selects an FNN from that pool. The family-competition with different mutation operator is repeated until the best FNN is found. Many others found that the adaptation in FNN nodes by one of the methods mentioned above can improve FNN performance to some extent~\cite{eannAlberto2002,gaLeung2003,eannAugusteijn2004,eannAbraham2007}. 

The third aspect of node optimization is to design a node as a model itself. Such modification leads to variate of neural network paradigms such as \textit{polynomial neural network}~\cite{oh2002design,andoni2014learning}, where the nodes are designed to as a polynomial function based on inputs to the nodes. Similarly, the nodes of a \textit{GMDH neural network} is designed as an Ivakhnenko polynomial~\cite{puig2007gmdh}; the nodes of a \textit{complex value neural network} or \textit{multivalued neural network} is designed with a complex value activation functions~\cite{hirose2006complex}; the node of \textit{spiking neural networks} has specific behavior, in which a node signal is propagated to another node only if the intrinsic quality of neural activation value is above a defined threshold~\cite{sporea2013supervised}; the nodes of \textit{fuzzy neural network} paradigm is designed using the concepts of fuzzy theory~\cite{fuller2013introduction}; the node and the architecture of the \textit{Quantum neural network} are inspired by the quantum computing~\cite{da2016quantum,narayanan2000quantum,kouda2005qubit,li2013hybrid}. In all such methods, metaheuristics have a significant role in the optimization.

\subsubsection{Learning algorithm optimization}
The initial thought of learning algorithm optimization is the optimization of its parameters. For example, the optimization the learning rate and the mutation factor parameters of BP by applying some metaheuristics~\cite{eannBelew1990evolvingnetworks}. To optimize the parameters of an FNN learning algorithm, its parameters (e.g., BP parameters) and learning rules are encoded onto a genotype~\cite{eannHarp1989,eannBaxter1993evolution}. However, formulating BP parameter such as learning rule, which is a dynamic concept, into a static chromosome is disadvantageous~\cite{eannChalmers1990evolution}. Hence, a genetic coding for four components (current weight, activation function of the incoming node and outgoing nodes, input) local to weight in an FNN can encode~\cite{eannChalmers1990evolution}. Moreover, assuming that each node in a network uses same learning rule, an evolution in learning was proposed in~\cite{eannKim1996fast}, where weights optimization related to a particular node depended only on the input/output at that node. Evolution in learning rule can be described as~\cite{eannXinYao1999}:
\begin{equation}
\label{eq_genLearning}
\Delta \text{\bf{w}}^t = \sum\limits_{k = 1}^{n}\sum\limits_{i_1, i_2, \ldots, i_k = 1}^{n} \left( \theta_{i_1, i_2, \ldots, i_k} \prod\limits_{j=1}^{k}w_{ij}^{t-1} \right)
\end{equation}
where $t$ is the iteration, $\Delta \text{\bf{w}}^t$ is weight change, $w_1, w_2, \ldots, w_n$ are weights associated with a node, $\theta_i$ is real-valued coefficient that is determined by using EAs. However, learning rule~\eqref{eq_genLearning} is impractical because of it required huge computation time. Hence, bio-inspired algorithms may be employed for determining the coefficient in~\eqref{eq_genLearning}. 

\subsubsection{Combination of FNN components optimization}
\begin{figure}
	\begin{center}
		\includegraphics[scale=0.6]{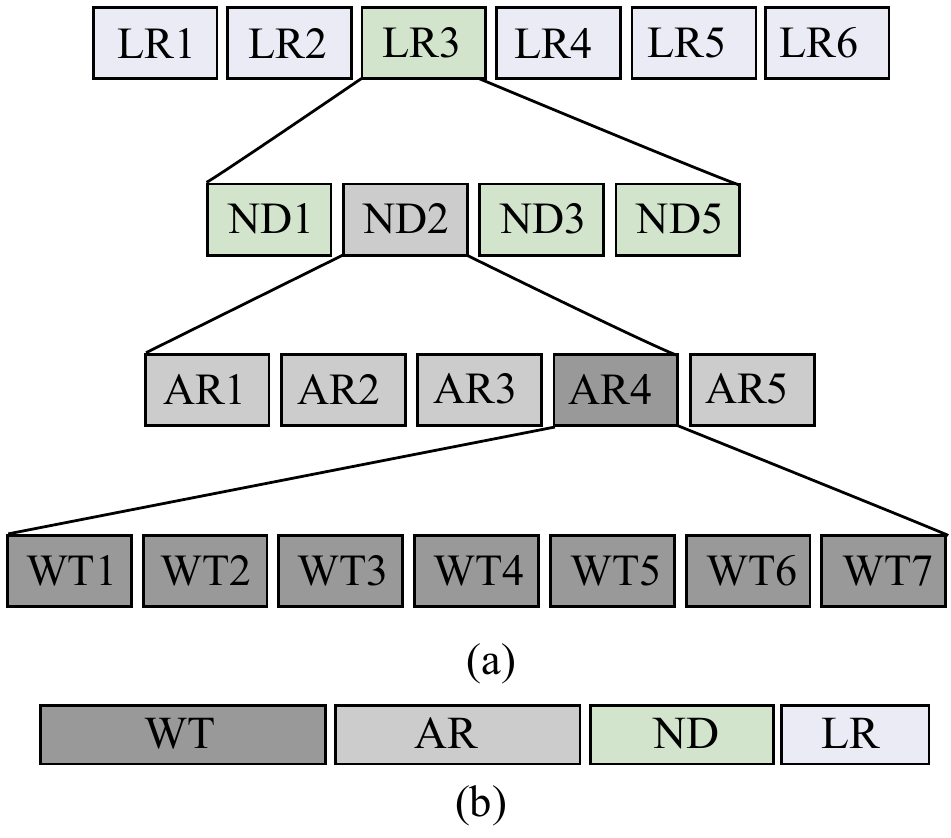}
	\end{center}
	\caption{{Meta-learning scheme (a), where LR is learning parameter, ND is activation function, AR is architecture, and WT is weight~\cite{eannAbraham2004}. Combined chromosome structure (b)}}
	\label{fig_FNNgenEncoding}
\end{figure}
\label{sussub:cooprative}
Fig.~\ref{fig_mhFNN} is an impressive representation of the most of FNN optimization combinations, where the genetic representation of the combination of FNN components can be represented in Fig.~\ref{fig_FNNgenEncoding}, which refers to a hierarchy of combination, called meta-learning scheme. In the meta-learning scheme, top down or bottom up optimization approach, which means, weights to learning rule and learning rules to weight optimization can be adopted~\cite{eannAbraham2004}. However, it resembles one-by-one learning  scheme. Hence, the advantageous approach is to represent each component of FNN side-by-side onto a genetic vector for optimization, which indicates the confluence of all components of FNN as indicated by area ``a8" in Fig.~\ref{fig_mhFNN}. 

Yao~\cite{eannXinYao1993,eannXinYao1999} summarized all such form of adaptation in {\textbf{evolutionary artificial neural network}} (EANN), which is a special class of artificial neural network, where in addition to learning; evolution is another fundamental form of adaptation. Infact a paradigm, called \textit{Neuroevolution} that accommodates adaptive learning all or some components of FNN in some intuitive ways by applying EAs. For examples, generalized acquisition of recurrent links (GNARL)~\cite{angeline1994evolutionary}, evolutionary programming net (EPNet)~\cite{eannXinYao1997}, neuroevolution of augmenting topologies (NEAT)~\cite{eannStanley2002evolving}, hypercube-based neuroevolution of augmenting topologies (HyperNEAT)~\cite{gauci2007generating}, evolutionary acquisition of neural topologies (EANT2)~\cite{kassahun2005efficient}, and heterogeneous flexible neural tree (HFNT)~\cite{ojha2016ensemble} optimizes both FNN structure and parameters (weights) using some direct or indirect encoding methods. Moreover, several other paradigms and methods proposed in the past for the simultaneous optimization of FNN components are described as follows.

A {\textbf{structured genetic algorithm}} was proposed in~\cite{gaDasgupta1992}, which simultaneously optimized both architectures and weights. It was found that the simultaneous optimization of both weight and architecture lead to a better generalization ~\cite{gaKitano1994,eannManiezzo1994,eannGirosi1995regularization,gaArifovic2001}. Considering permutation\footnote{Permutation problem occurs when using  traditional crossover operator, where a population has traditional genetic representation of FNN architecture.} problem in a GA, EP-based mutation mechanism for evolving FNN architecture was proposed in~\cite{eannXinYao1997} is known as EPNet.

A {\textbf{neuroevolution of augmenting topologies}} (NEAT) introduced in~\cite{eannStanley2002evolving} was a GA-based evolution of an FNN phenotype as a whole, in which a special mutation and crossover operator were defined for manipulating nodes and connections of FNN. Specifically, the linear network information FNN weights, nodes, and connection information were encoded using genetic encoding. The proposed NEAT was evaluated over several applications, and its performance was found outperforming static FNN topology.  

A {\textbf{virtual subpopulation}} approach was proposed in~\cite{gaSalajegheh2005} for the optimization of FNN using EAs. Later, while indicating a permutation problem, crossover operator as a combinatorial optimization problem was proposed in which each hidden node was considered as a subnetwork and a complete network was evolved using the evolution of several subnetworks~\cite{gaPedrajas2006}. Additionally, GA-based and SA-based crossover operators were applied to generate an offspring (new individual subnetwork). To maintain diversity in population, two mutation operators such as BP-mutation and random-mutation were proposed. In BP-mutation, few iterations of BP algorithm were applied to update weights of the subnetwork, and in random mutation, weights of subnetwork were randomly replaced with new weights. Hence, a \textit{coevolution} of FNN weights and architecture was proposed that evolved FNN with the \textit{cooperation} of the individuals of a subnetwork population.

A {\textbf{cooperative coevolution neural network}} process---inspired by virtual subpopulation approach~\cite{gaSalajegheh2005}--- was proposed in~\cite{eannMoriarty1997forming}, which was a symbiotic, adaptive neuroevolution (SANE) algorithm for constructing FNN in a dynamic environment. Unlike conventional evolutionary approach, which uses a population of FNNs, SANE uses a population of nodes, where each node establishes connections with the other nodes to form a complete network.

Two reasons of better performance of SANE over conventional and stand-alone metaheuristics were suggested. First, since SANE consider the nodes as functional components of the FNN, it accurately searches and evaluates nodes as genetic building blocks. Second, since a node alone cannot perform well and evolutionary process evolves different types of nodes, SANE was able maintains diversity in the population. Later, the concept of SANE was extended, in which the selection of several individuals from a population of hidden nodes was combined in a various permutation in order to form several complete networks, i.e., evolution in hidden nodes led to an evolution of the complete network~\cite{annEmoGarciaPedrajas2003cov}.

A concept of {\textbf{sparse neural trees}}, in which GP for evolving network structure and GA for parameter optimization was proposed in~\cite{zhang1997evolutionary}. Similarly, a \textit{flexible neural tree} (FNT) concept, where GP was used for the adaptation in network structure and SA for the optimization of the parameters (including parameters of activation function) was proposed in~\cite{eannChen2004125,eannChen2006}. FNT is a tree-like model where adaptation in all components of is equally important. Moreover, its components adaptation may take many forms (Fig.~\ref{fig_mhFNN}). Hence, a beta basis function---which has several controlling parameters, such as shape, size, and center---was used at non-leaf nodes of an FNT~\cite{eannBouaziz2014universal}. It was observed that embedding beta-basis function at FNT nodes has advantages over other two parametric activation function. A parallel evolution of FNT using MPI programming and GPU programming respectively were proposed in~\cite{eannPeng2011parallel} and in~\cite{eannWang2012modeling}.

A slightly different direction of FNN modification and improvement study can be seen as the study of {\textbf{quantum neural network}} (QNN). At the first place, the QNN as a \textit{quantum perceptron} was proposed by Lewestein~\cite{lewenstein1994quantum}, where instead of classical weights, a unitary operator was used to map inputs to an output. The study in QNN encompasses the development of quantum weights, quantum neurons, quantum network, and quantum learning~\cite{narayanan2000quantum}. 

The design/algorithm of quantum network was thought as an algorithm that can find the control parameters for a coupled qubit system~\cite{gershenfeld1998quantum} as it appears in quantum computing. A comprehensive quantum inspired neural network is presented in~\cite{menneer1995quantum}, where two categories of inspiration were drawn: strongly and weakly quantum inspired FNN. In strongly inspired QNN, each pattern in a training set was considered as a particle which is processed by a number of FNNs in different universes. Such process was compared with the electrons or photons passing through many slits simultaneously. Whereas, in weakly inspired QNN, each training pattern (though as a particle) was in its own universe. Moreover, there were various QNN models proposed in the past by 1) Behrman et al.~\cite{behrman1996quantum}, 2) Chrisley~\cite{chrisley1997learning}, 3) Menneer~\cite{menneer1995quantum}, and Ventura~\cite{ventura1998artificial}. A detailed description of these QNN models can be found in~\cite{rudolph2011heuristic}.
  
\subsection{Comments on metaheuristics approaches}
It is indeed can be concluded that metaheuristic algorithms have provided various dimension to the optimization of the FNNs. It has opened several ways such that a generalized FNN can be obtained. Especially the architecture simplification which is directly related to the generalization of an FNN can easily be achieved through the evolving FNN together with its other components. However, the primary disadvantage of using metaheuristic algorithms is the training time consumption. Since the metaheuristic algorithms use a population (many solution candidates) during optimization, the time consumption becomes directly proportional to the number of candidates in a population. 

It can also be argued that both conventional and metaheuristic based FNN training take by far more training time than the \textit{extreme learning machine} (ELM)~\cite{huang2006extreme}. ELM is a three-layered FNN architecture whose weights between input and hidden layer are randomly assigned and never updated. Additionally, the weights between hidden and output layer are updated in a single step using some least square estimation. Hence, extremely less time required for the learning of the FNN. 

It is stated in NFL theorem~\cite{mhWolpert1996lack} that it is difficult to find a metaheuristic algorithm that solves all class of problems. Hence, a metaheuristic algorithm may find difficulty in optimizing the FNN that has been formulated for solving some specific problem (input patterns). Additionally, it is not theoretically possible to understand or determine that how fast a metaheuristics algorithm will converge or finds a satisfactory solution. The only way to determine a metaheuristic algorithm's convergence is by its empirical evaluations. Moreover, since each metaheuristic applies some specific heuristic, it is difficult to select one metaheuristic as the best metaheuristic at an instance for a problem. It is only possible to select a metaheuristic by empirically comparing the convergence speeds and trained FNN performances. 

\section{Multiobjective metaheuristic approaches}
\label{sec_fnnEMO}
\label{subsec_formulation}
Multiobjective optimization procedure involves in optimizing two or more objectives functions, simultaneously. Multiobjective algorithms are efficient methods for evaluating Pareto-optimal solutions for multiobjective problems. Since optimizing training error cannot provide generalization alone, FNN optimization is viewed from the multiobjective perspective.

Let us first investigate: \textit{why the multiobjective framework is needed for FNN optimization}, \textit{what are the objective functions required for framing FNN as a multiobjective problem}, and \textit{how the objective functions can be framed into multiobjective optimization}. Answers to these question lie in the following discussion.

First, a cost function~\eqref{eq_mse} or any equivalent function is the foremost necessity for the supervising training of FNN. 

Second, the generalization of FNN is an essential aspect of its optimization. One approach is to use validation error on cross-validation data because an FNN with low training error may not perform well on unseen (test) data unless FNN is generalized. Moreover, minimization of  generalization error is essential than the minimization of training error.

Another approach is to add a regularization term to the training error to avoid overfitting. Additionally, minimizing network complexity leads to a better generalization~\cite{jin2005evolutionary}. Hence, generalization can be achieved by adding a complexity indicator term to training error, i.e., the generalization by minimizing training error and simplifying network complexity.  

Third, reducing input-feature---when a problem is available with a huge input dimension (feature)---can lead to a better generalization. However, input dimension reduction  and training error reduction are two contradictory objectives. 

Finally, the conclusion is, the training error~\eqref{eq_mse} or equivalent cost function $ c_f $ needs to be optimized with one or more additional objectives to achieve generalization, which is why multiobjective framework for optimizing FNN are used. 

Let us say that training error~\eqref{eq_mse} is $c_{trn}$, and an additional objective is $c_{add}$. So, a generalized error $c_{gen}$ may be computed by adding an objective to training error as:
\begin{equation}
\label{eq_genError}
c_{gen} =  c_{trn} + \lambda c_{add},
\end{equation}
where $\lambda > 0 $ is a hyperparameter that controls the strength of additional objective $c_{add}$. The validation error term $c_{cv}$, regularization term $c_{reg}$, or network complexity $c_{net}$ or a combination of all can be considered as an additional objective $c_{add}$ in~\eqref{eq_genError}. The regularization term $c_{reg}$ is the weight decay or norm of weight vector $\text{\bf{w}}$ as:
\begin{equation}
\label{eq_normWeight}
c_{reg} = \frac{1}{2} \sum\limits_i^n w_i^2 = \frac{1}{2}\parallel \text{\bf{w}} \parallel^2.
\end{equation}
Similarly, a validation error $c_{cv}$ is usually computed using~\eqref{eq_mse} on a cross-validation data. On the other hand, the network complexity $c_{net}$ is computed as: 
\begin{equation}
\label{eq_netComplexity}
c_{net} = \sum\limits_i^{z}\sum\limits_j^{z} c_{ij},
\end{equation}
where $z$ is the number of nodes in the network $c$ (Fig.~\ref{fig_ArcDirect}), or any user-defined function can also be used for evaluating network complexity, e.g., the number of nodes, the number of connections, etc. 

However, the generalization objective of the form~\eqref{eq_genError} is a scalarized objective that has two disadvantages~\cite{das1997closer}. First, determining an appropriate hyperparameter $ \lambda $ that controls the contradicting objectives. Hence, the generalization ability of the produced model by using~\eqref{eq_genError} will be a mystery. Second, the objective~\eqref{eq_genError} leads to a single best model that tells nothing about how contradicting objectives were handled. In other words, no single solution exists that may satisfy both objectives. Therefore, generalization error~\eqref{eq_genError} need to be formulated into a multiobjective form: $ \min \{ c_{trn}, c_{reg}, c_{cv}, \ldots \}$, i.e., a multiobjective optimization needs to be performed as:
\begin{flushleft}
 $\mbox{minimize } \{c_1(\text{\bf{w}}), c_2(\text{\bf{w}}), \ldots, c_m(\text{\bf{w}})\}$\\
 $\mbox{subject to } \text{\bf{w}} \in S,  \nonumber $\\
\end{flushleft}
\noindent where $m \ge 2$ is the number of objective functions $c_i: \mathbb{R}^n \to  \mathbb{R}_{\ge 0}$. The vector 
of objective functions is denoted by $ \text{\bf{c}} = \langle c_1(\text{\bf{w}}), c_2(\text{\bf{w}}), \ldots, c_m(\text{\bf{w}}) \rangle $. The decision (variable) vectors $\text{\bf{w}} = \langle w_1, w_2,\ldots, w_n \rangle$
belong to the set $S \subset  \mathbb{R}^n$, which is a subset of the decision variable space $ \mathbb{R}^n$. 
The word ``minimize" indicates the minimization all the objective functions simultaneously. 

A nondominated solution is one in which no one objective function can be improved without a simultaneous detriment to at least one of the other objectives of the solution. The nondominated solution is also known as a Pareto-optimal solution.
\begin{mydef}
	\label{def_Pareto_dom}
	Pareto-dominance - A solution  $ \text{\bf{w}}_1 $ is said to dominate a solution $ \text{\bf{w}}_2 $ if $ \forall i = 1,2,\ldots,m $, $ c_i(\text{\bf{w}}_1) \le c_i(\text{\bf{w}}_2) $, and there exists $ j \in \left\lbrace 1,2,\ldots,m\right\rbrace  $ such that $ c_j(\text{\bf{w}}_1) < c_j(\text{\bf{w}}_2) $.
\end{mydef}
\begin{mydef}
	Pareto-optimal - A solution $ \text{\bf{w}}_1 $ is called \textit{Pareto-optimal} if there does not exist any other solution that dominates it. A set \textit{Pareto-optimal} solution is referred to as a Pareto-front.  
\end{mydef}

Now, a multiobjective algorithm must provide a homogeneous distribution of a population along Pareto front and improve solutions along successive generations~\cite{emoGaspar2005multi}. Hence, three basic operators can be used~\cite{emoGaspar2005multi}. 
1) \textit{Fitness assignment} to guide a population in the direction of Pareto-front using robust and efficient multiobjective selection method.
2) \textit{Density estimation} to maintain solutions distributed over entire Pareto-front using operators that take account of the solution's proximity.
3) \textit{Archiving} to prevent degradation in fitness during successive generations by maintaining an external population to preserve  the best solutions and for periodic input to the main population. A detailed  survey of multiobjective algorithms is available in ~\cite{emoCoello1999survey,zhou2011multiobjective}. Now, based on the above discussion on the cost functions and generalization conditions, the multiobjective for FNN optimization can be categorized as \textit{non-Pareto based multiobjective} approach and \textit{Pareto-based multiobjective} approach.

\subsection{Non-Pareto based multiobjective approaches}
In non-Pareto based multiobjective approach, the objective functions are aggregated as mentioned in~\eqref{eq_genError} or by some other means. For example, in~\cite{annEmoTeixeira2000}, authors proposed to add a regularization term $c_0 - c_{reg} $ to training error $ c_0 - c_{trn} $, where $ c_0 $ is the origin of the two objectives. To obtain an efficient solution, they designed a vector $ \textbf{vc} $ of \textbf{scalar objectives} by varying the hyperparameter $\lambda $ from 0 to 1. Hence, training FNN for each scalar objective in vector $ \textbf{vc} $, a Pareto set was obtained, and then, it was possible to select the best solution from the Pareto front. However, this was an expensive approach, which does not use any Pareto-based multiobjective algorithm to compute Pareto set; rather, an ellipsoid method~\cite{bland1981ellipsoid} was applied to train FNN for each scalar objective $ vc_i \in \textbf{vc} $ sequentially. 

Similarly, in~\cite{annEmoCosta2003training}, to achieve generalization, authors proposed a \textbf{sliding mode control BP} algorithm for the multiobjective treatment to FNN objectives $c_{trn}$ and $c_{reg}$. The optimization trajectory of the 2D space of the objectives $c_{trn}$ and $c_{reg}$ was controlled by modifying BP weight update rules using two sliding surface control indicators each belongs to the mentioned objectives, respectively

Multiobjective treatment to FNN was also offered by using improvising metaheuristics itself such as a \textit{predator-prey} algorithm was proposed in ~\cite{annEmoPettersson2007}. To get a generalized network, the predator-prey algorithm used a family of the randomly generated population of sparse neural networks, called \textit{pray population} and an externally induced family of \textit{predators population} whose job was to prune preys populations based on the objectives $c_{trn}$ and $c_{net}$ was also generated. Similarly, a hybrid multiobjective approach, where a geometrical measure based on singular-value-decomposition for estimating a necessary number of nodes in a network  was proposed in~\cite{annEmoGoh2008}. 

Additionally, a micro-hybrid genetic algorithm was introduced to fine-tuning the network performance. A hybrid algorithm, which uses GA for evolving FNN and uses PSO, BP, and LM for fine-tuning the evolved FNN was proposed in~\cite{annEmoAlmeida2010}. In the proposed hybrid algorithm, several objectives function such as training error $ c_{trn} $, validation error $c_{cv}$, number of hidden layers $c_{hid}$, number of nodes $c_{node}$, and activation function $c_{fun}$ were aggregated as:
\begin{equation}
\label{eq_multObj}
c_{net} = \alpha c_{trn} + \beta c_{cv} + \gamma c_{hid} + \delta c_{node} + \theta c_{fun},
\end{equation}
where, $\alpha$, $\beta$, $\gamma$, $\delta$, and $\theta$ were controlling parameters. Hence, multiple objectives were optimized simultaneously. 

As mentioned above in Section~\ref{sec_fnnEMO}, the \textbf{aggregating objective} function has disadvantages in obtaining the best generalized solution. It is evident from~\eqref{eq_multObj} that determining hyperparameters for controlling objective function is a challenging task. Therefore, Pareto-based multiobjective is an efficient choice for the multiobjective treatment of FNNs.
  
\subsection{Pareto based multiobjective approaches}
The advantages of applying Pareto-based learning is thoroughly explained and compared with a single and scalerized objective in~\cite{jin2008pareto}. For example, a \textbf{nondominated sorting genetic algorithm version~II} (NSGA-II)~\cite{emoDeb2000nsgaII} when used for optimizing objectives $c_{trn}$ and $c_{net}$ offers a Pareto set by  optimizing both objectives simultaneously using a nondominated sorting method as defined in Definition~\ref{def_Pareto_dom}. Hence, NSGA-II can be applied to obtained a regularized network by optimizing the objectives $c_{trn}$ and $c_{reg}$~\cite{annEmoJin2004}. 

Similarly, \textbf{Pareto differential evolution} (PDE) algorithm and its variant self-adaptive PDE algorithm was applied to optimize objectives $c_{trn}$ and $c_{net}$ simultaneously that offered a Pareto-set, from which the best solution was picked-up according to network complexity and approximation error examination~\cite{annEmoAbbass2003speeding,annEmoAbbass2002self}. Simultaneous optimization of the objectives $c_{trn}$ and $c_{net}$ were also addressed using \textbf{multiobjective PSO} to generalize FNN performance~\cite{emoYusiong2006training}. 

For an image classification problem, authors in~\cite{annEmoWIEGAND2004}, pointed out two crucial points: the classification speed and the classification accuracy $ c_{acc} $. The classification speed was then related to the network complexity (number of hidden neurons) $ c_{net} $. The proposed trade-offs between classification speed and classification accuracy were addressed using NSGA-II. 

Similarly, in~\cite{annEmoRoth2006}, authors studied three methods for image classification problem: linear aggregating (LA), NSGA-II with deterministic selection (DM), and NSGA-II with tournament selection (LM). They proposed to optimize network complexity $c_{net}$ and accuracy $ c_{acc} $. Moreover, they combined regularization term $ c_{reg} $ with accuracy $c_{acc}$ and proposed an adaptive strategic for designing network topology using reproduction operators for both hidden layer and input layer. The hidden layer operators were add-connection, delete-connection, add-node, and delete-node. The receptive (input) layer had the following operators: add-connection, delete-connection, add-node, and delete-node. Interestingly, they observed that DM and LM performed better than LA, i.e., Pareto-based multiobjective algorithms performed better than that of the scalerized objectives. Such ability of the Pareto-based treatment to FNN to obtain general FNN was exploited by several researchers for solving many real-life applications~\cite{emoJin2005evolutionary,annEmoFurtuna2011,eannZavoianu2013,psoKarpat2007}.

Further, the \textbf{coevolution FNN} concept~\cite{annEmoGarciaPedrajas2003cov,annEmoGarciaPedrajas2002sym} was extended in~\cite{annEmoGarciaPedrajas2002} under the multiobjective framework, by using \textit{subnetwork} and \textit{network} concepts. A subnetwork was a collection of nodes, i.e., a subnetwork was considered as a hidden node for a network. Therefore, a network was a collection of subnetworks. So, a population $P_1$ of subnetwork, which was evolved separately using NSGA-II was used to construct a population $P_2$ of networks. Then, NSGA-II was again applied to evolve population $P_2$. Interestingly, authors defined separate objectives for population $P_1$ (subnetworks objectives) and $P_2$ (networks objectives) so that the functional diversity in both network and subnetwork can be maintained. Additionally, some metrics (objectives) for measuring network and subnetwork functional diversities were defined. The objective of subnetworks were \textit{differences} (for maintaining functional diversity of subnetwork), \textit{substitution} (to replace poor candidates by better candidates), and \textit{complexity} (for counting  the number of connection, nodes, and sum of all weights). Therefore, they coevolved overall network with the cooperation of subnetwork that evolves together with the whole network to get a general solution to a problem. 

Apart from the discussed objective in this section, some interesting dimensions in multiobjective treatment to FNN can be noted in~\cite{annEmoGiustolisi2006optimal}, in which authors proposed to apply NSGA-II for the simultaneous optimization of three objectives: \textit{input-dimension}, training error, and network complexity. Hence, an optimized a network that performs well on the minimal set of input dimension was obtained. Similarly, in~\cite{mhCruz2010memeticPareto,classFernandez2010}, authors used a Pareto-based memetic algorithm approach for combining PDE and Rprop algorithms to minimize objective pairs \textit{true classification rate} and \textit{minimum sensitivity} (miss-classification rate), simultaneously. 

As a result of metaheuristic or multiobjective metaheuristic treatment, a set of FNN network is obtained and selecting the best FNN from that set is a difficult task. Since selecting a single best FNN from may not offer a generalized solution and the residual error can still be remaining many problems when selection single best FNN~\cite{ensmblSejnowski1987parallel}, then an ensemble of a set of FNNs is recommended. 

\section{Ensemble of feedforward neural networks}
\label{sec_ensemble}
Metaheuristics optimization of FNN leads to a final population that contains many solutions close to the best solution. Moreover, the solution in the final population are divers in the following sense: 1) parametric (each FNNs have different sets of weights); structural (each FNNs have different network configurations); and 3) training set (each FNNs are trained on different parts of a training set). Hence, a collective decision (ensemble) of $ l $ many diverse candidates selected from a final population may offer desired generalization~\cite{ensmblXinYao1998}. The literature that explains \textit{how to construct diverse FNNs} and \textit{how to combine decisions of diverse FNNs} are summarized as follows.      

The very basic idea is to apply single-solution based algorithms on $ l $ many FNNs to get $ l $ many diverse solutions~\cite{ensmblHansen1990}. The decision of $ l $ many candidates which were created either by single solution-based metaheuristics, or by  population-based metaheuristics, or by any other means are combined using the following methods~\cite{polikar2006ensemble,ensmblXinYao1996}: 1) majority voting method (for classification problems); 2) arithmetic mean (for regression problem); 3) rank-based linear combination; 4) linear combination by using \emph{recursive least square}~\cite{ensamblDavis1985stochastic} (to minimize weighted least squares error so that redundant individuals are eliminated); 5) evolutionary weighted mean or majority voting (metaheuristic to determine impact of an FNN in ensemble); and 6) entropy-based method for combining FNNs in ensemble (assigning entropy to FNNs during the learning process)~\cite{ensmblZhao2011}.

Since population-based metaheuristics lead to an optimized final population, it is advantageous to use the final population for making ensemble~\cite{ensmblXinYao1998}. However, there are two fundamental problems with it~\cite{ensmblLiu1999ensemble}: 1) determining ensemble size, 2) how to maintain diversity in the population. Hence, a \textbf{negative correlation learning} (NCL) algorithm that optimized and combined individual FNNs in an ensemble during learning process was proposed in ~\cite{ensmblLiu1999ensemble}. NCL optimized all individual FNNs simultaneously and interactively by adding a correlation penalty terms to the cost functions. Moreover, NCL produced negatively correlated and specialized FNNs by using co-operation among each FNNs of a population~\cite{psoZhengQin2005}. 

To determine the size of ensemble automatically, EA-based ensemble procedure was laid down in which NCL was applied during networks training. Moreover, different FNNs were allowed learn different parts of training data and the best (according to fitness) were selected for ensemble~\cite{ensmblXinYao2000}. Additionally, a \textbf{constructive-cooperative-neural-network-ensemble} was proposed in~\cite{ensmblIslam2003constructive} that determined ensemble size by focusing on accuracy and diversity during a constructive, cooperative procedure~\cite{ensmblYao2008evolving}.

However, mere training fitness based selection of candidates for the ensemble is insufficient because it does not tell much about candidates role/influence in the ensemble. This problem was addressed in a \textbf{GA-based selective ensemble} method~\cite{ensmblZhou2002}, which selects a subset of  the population and determine the strength of selected candidates using GA. It was also shown that the ensemble of a subset of the population was found performing better than that of the whole population~\cite{ensmblZhou2002}. The effectiveness such GA-based selection was found efficient than the traditional ensemble methods:~\textit{bagging}~\cite{ensmblBreiman1996bagging} and \textit{boosting}~\cite{ensmblSchapire1990strength}. 

It is beneficial to partition/fracture training data and allows different FNN in the population to learn various  parts of training data~\cite{ensmblBreiman1996bagging,ensmblSchapire1990strength}. An evidence of such was examined in~\cite{ensmblBakker2003,ensmblChen2010}, where it was found that the ensemble of a few FNNs that was trained using \textit{bootstrapping} performs better than that of an ensemble of a larger number of FNNs. Similarly, the efficiency of using distinct training sets for optimizing different FNNs was proved when a \textbf{class-switching ensembles} approach proposed in~\cite{ensmblMartínez2008} and were compared with bagging and boosting methods.  

At one hand bootstrapping method allows FNN to learn different training samples. On the contrary, a \textbf{clustering-and-coevolution} approach for constructing neural network ensembles proposed in~\cite{ensmblMinku2008} partition the input space using a clustering method to reduced number of input nodes of FNNs. Hence, in the ensemble, diverse FNNs (different FNNs were specialized in various regions of input space) were created. Moreover, it reduced run time of learning the process by coevolving (divide-and-conquer method) different FNNs using cooperation between FNNs. Such method improves diversity and accuracy of an ensemble system~\cite{ensmblMinku2008}. 

Similarly, a method was suggested in~\cite{ensmblKim2008} for generating diverse evolutionary FNNs using a fitness-sharing method---a fitness sharing method shares resources if the distance between the individuals is smaller than the predefined sharing radius. Specifically, authors proposed a speciation based evolutionary neural ensemble method for constructing ensemble by combining FNNs using a knowledge space method. On the other hand, a progressive interactive training scheme called  a \textbf{sequential-neural-network-ensemble-learning} method, which trained FNNs one-by-one by interaction from a central buffer of FNNs was proposed in~\cite{ensmblAkhand2009}. 

Both diversity and accuracy is a crucial aspect in construing ensemble of FNNs~\cite{polikar2006ensemble}. However, accuracy and diversity are contradictory to each other, so, a multiobjective approach may be applied to evolve FNN population by maintaining accuracy and diversity simultaneously~\cite{annEmoXinYaoChandra2006}. For this purpose, \textbf{multiobjective regularized negative correlation learning} that maximized performance and maximized the negative correlation between individuals in population was found efficient~\cite{annEmoXinYao2010}. 

\section{Challenges and future scopes} 
\label{sec_challanges}
The effectiveness of FNN training primarily depends on \textit{data quality}, which is governed by the following \textit{data quality assurance} parameters: accuracy, reliability, timeliness, relevance, completeness, currency, consistency, flexibility, and precision~\cite{dataQtyWand1996anchoring,dataQtyPipino2002data}. Usually, \textit{data cleaning} is a major step  before modeling~\cite{dataCleanHernandez1998real}. Therefore, training of the FNN remains always sensitive to the data cleaning process and it poses a significant challenge to adapt some mechanism in training process such the sensitivity towards data-clean may be reduced. Additionally, one problem related to data-driven modeling (FNN learning) is the data itself which can be insufficient, imbalanced, incomplete, high-dimensional, or abundant. 

For the case \textbf{insufficient data}, usually the input hyperspace is exploited to generate virtual samples to fill the sparse area of the hyperspace, and by monitoring FNN performance on the virtually generated samples~\cite{cho1997virtual}. The second approach exploits the dynamics of EAs in conjunction with FNNs to obtain new samples~\cite{eannZhang1991neural}. However, this area is still much to explore, where some open questions such as how efficiently FNNs can be trained with virtually generate data to mitigate the insufficiency. On the other hand, research in the area of imbalance dataset is continued to interest researcher~\cite{mazurowski2008training}.

The present era of data analysis is what we call \textit{big data}, i.e., we need to deal not only with \textit{high-dimensional data} but also with the \textit{variety data} and \textbf{stream data}~\cite{zikopoulos2011understanding}. High-dimensional data, such as gene expression data, speech processing, natural language processing, social-network-data, etc., poses significant challenges. Such challenge is to some extent addressed by \textit{deep learning} paradigms that allow the arrangement several units/layers of FNNs (or any other model) in a hierarchical manner to process and understand insights of such high-dimensional data~\cite{deepHinton2012,deepHinton2006fast}. High-dimensional data can also be managed/reduced by encoding or decoding methods and using FNNs training~\cite{hinton2006reducing}. Therefore, FNNs has a greater role in feature reduction. 

In a \textbf{non-stationary environment}, such as stock-price market, weather forecasting, etc., data comes in the stream, i.e., data comes in sequential order, and traditionally, re-training based mechanics for dynamic learning (online learning) of FNN is the basic option~\cite{saad2009line}. However, it is still an open problem to design strategies for the dynamic training of FNN.

Apart from the crucial aspects that \textit{how to manage non-stationary data}, the aspect that how to handle \textbf{multi-view (heterogeneity)} of data is an additional challenge. The quest of developing a model that can stand robust and efficient for the non-stationary data caused by the time-dependent process of data generation, and can accommodate new knowledge (newly generated data sample) is a significant topic in machine learning research~\cite{ditzler2015learning}.  On the other hand, integration of data or of the models for that matter for the heterogeneous data generated or gathered from different instruments and data-generation processes is a significant research problem~\cite{ritchie2015methods,pavlidis2001gene}.

Moreover, present era, the \textbf{fourth industrial revolution}, is of \textit{Internet of Things} (IoT)~\cite{prisecaru2016challenges}. In IoT, sophisticated technologies such as \textit{smartphone} and \textit{smartwear} provide several forms of data, e.g., \textit{human activity recognition}~\cite{kim2010human}. Additionally, it demands application to be simple. Hence, FNN models which when aims to such technologies needs to be less complex. Therefore, FNN architecture simplification or model's complexity reduction is a challenging task. Such problem can be addressed through the integration of FNN with statistical methods like the one usually done with \textit{hidden Markov model}\cite{trentin2001survey}. Therefore, such kind of modification to network architecture and specialized node design may lead to different paradigms of FNN that may solve various real-world complex problems.

\section{Conclusions}
\label{sec_con}
Feedforward neural network (FNN) is used  for solving a wide range of real-world problems, which is why researcher investigated many techniques/methods for optimizing and generalizing FNN. Specifically, metaheuristics allow us to innovate and improvise methods for optimizing FNN that in turn address its local minima and generalization problems. 

Initially, only gradient based linear approximation and quadratic approximation methods for optimizing FNNs were employed to train FNN. These conventional algorithms (backpropagation, Quickpro, Rprop, conjugate gradient, etc.) are local search algorithms that exploit current solution to generate a new solution; however, they lack in exploration ability, therefore, usually, finds local minima of an optimization problem. 

Unlike conventional approaches, metaheuristics (e.g., genetic algorithm, particle swarm optimization, ant colony optimization, etc.) are good at both exploitation and exploration and can address simultaneous adaptation in each component of FNN. However, no single method can solve all kinds of problem. So, we need to improvise, adapt, and construct hybrid methods for optimizing FNN. Therefore, several dynamic designs of FNN are reported in the literature: EPNet (an adaptive method of FNN architecture optimization), neuro-evolution of augmenting topologies, flexible neural tree, cooperative coevolution neural network, etc., are among them. Hence, there is a wide spectrum of FNN optimization/adaptation is possible with metaheuristic treatment to FNNs (Fig.~\ref{fig_mhFNN}) in which the fundamental aspect is the formulation of FNN (phenotype) to vectored form (genotype) or any other form of mechanism for manipulation of FNN components. 

Since there are many components to be manipulated by means of metaheuristic strategies and the availability of the fact that FNN generalization ability depends on the optimization its all the components, multiobjective treatment to FNN were used. The multiobjective-based training allows an FNN to evolve with handling two or more FNN-related objectives, such as approximation error, network complexity, input dimension, etc. Moreover, the generalization ability of system can be easily improved by combining decision of many candidates of the system. Hence, an ensemble of FNNs by making use of the metaheuristic final population was proposed and the two crucial aspect accuracy and diversity of an ensemble were taken care during propose of evolving FNNs. 

It is evident from such aspects of FNN optimization that the future research will be able to bring the new paradigms of FNNs by applying or by the inspiration from the discussed methods in this article. Hence, that will overcome the data quality problem and will be handling new challenges of big data to cope-up with the new era information processing.

\section*{Acknowledgment}
Authors would like to thank the all the anonymous reviewers for the technical comments, which enhanced the contents of the preliminary version of this paper. This work was supported by the IPROCOM Marie Curie Initial Training Network, funded through the People Programme (Marie Curie Actions) of the European Union’s Seventh Framework Programme FP7/2007–2013/, under REA grant agreement number 316555.

\footnotesize
\renewcommand{\baselinestretch}{1} 
\bibliographystyle{IEEEtran}
\bibliography{survey_FNN_EAAI}

\end{document}